\documentclass[letterpaper, 10 pt, journal, twoside]{IEEEtran}

\IEEEoverridecommandlockouts                              

\usepackage{url}            
\usepackage{booktabs}       
\usepackage{amsfonts}       
\usepackage{nicefrac}       
\usepackage{microtype}      
\usepackage{xcolor}         
\usepackage{amssymb}  

\usepackage{graphicx}
\usepackage{amsmath}
\usepackage{amsthm}
\usepackage{booktabs}
\usepackage{algorithm2e}
\usepackage{setspace}
\usepackage{enumitem}
\usepackage{bbm}
\usepackage[]{caption}
\usepackage{subcaption}
\usepackage{tabularx}
\usepackage[rightcaption]{sidecap}
\usepackage{pifont}
\newcommand{\xmark}{\ding{55}}%
\usepackage{xcolor}
\usepackage{pifont}
\usepackage{color, colortbl}
\usepackage{siunitx}
\usepackage{wrapfig}
\usepackage{floatrow}
\newfloatcommand{capbtabbox}{table}[][\FBwidth]

\definecolor{brightgreen}{rgb}{0.4, 1.0, 0.0}
\definecolor{LightCyan}{rgb}{0.88,1,1}
\definecolor{LightGray}{gray}{0.9}
\definecolor{Gray}{gray}{0.9}
\newcommand*\colourcheck[1]{%
  \expandafter\newcommand\csname #1check\endcsname{\textcolor{#1}{\ding{52}}}%
}
\newcommand*\colourx[1]{%
  \expandafter\newcommand\csname #1x\endcsname{\textcolor{#1}{\ding{55}}}%
}
\usepackage{amsfonts}

\colourcheck{blue}
\colourcheck{green}
\colourcheck{red}
\colourcheck{brightgreen}
\colourcheck{black}
\colourcheck{gray}
\colourx{blue}
\colourx{green}
\colourx{red}
\colourx{black}
\colourx{gray}

\newcommand{\orangetext}[1]{\textcolor{orange}{#1}}

\newcommand{\quotes}[1]{``#1''}

\usepackage{hyperref}
\hypersetup{
    colorlinks=true,
    linkcolor=orange,
    filecolor=magenta,      
    urlcolor=orange,
    citecolor=orange,
}

\usepackage[capitalise, nameinlink]{cleveref}

\newcolumntype{a}{>{\columncolor{Gray}}c}

\usepackage{multirow}

\crefname{appendixfigure}{App. Fig.}{App. Figs.}
\crefname{appendixtable}{App. Table}{App. Tables}

\usepackage{mdframed}
\usepackage[frozencache,cachedir=.]{minted}

\usepackage[export]{adjustbox}
\usepackage{float}
\usepackage{stfloats}
\newcommand{\MethodName}{{MotIF}\xspace}
\newcommand{\DatasetName}{{MotIF-1K}\xspace}

\title{\LARGE \bf
MotIF: Motion Instruction Fine-tuning}

\author{
  Minyoung Hwang$^{1}$, Joey Hejna$^{2}$, Dorsa Sadigh$^{2}$, Yonatan Bisk$^{3}$ \vspace{-0.5cm}
  \thanks{This work has been submitted to the IEEE for possible publication. Copyright may be transferred without notice, after which this version may no longer be accessible.
  }
\thanks{$^{1}$Massachusetts Institute of Technology, $^{2}$Stanford University, and $^{3}$ Carnegie Mellon University. {\tt\footnotesize myhwang@mit.edu, jhejna@stanford.edu, dorsa@stanford.edu, ybisk@cs.cmu.edu}}
}

\begin{document}

\maketitle

\begin{abstract}
While success in many robotics tasks can be determined by only observing the final state and how it differs from the initial state -- e.g., if an apple is picked up -- many tasks require observing the full motion of the robot to correctly determine success. For example, brushing hair requires repeated strokes that correspond to the contours and type of hair. Prior works often use off-the-shelf vision-language models (VLMs) as success detectors; however, when success depends on the full trajectory, VLMs struggle to make correct judgments for two reasons. First, modern VLMs are trained only on single frames, and thus cannot capture changes over a full trajectory. Second, even if we provide state-of-the-art VLMs with an aggregate input of multiple frames, they still fail to correctly detect success due to a lack of robot data. Our key idea is to fine-tune VLMs using abstract representations that are able to capture trajectory-level information such as the path the robot takes by overlaying keypoint trajectories on the final image. We propose motion instruction fine-tuning (\MethodName), a method that fine-tunes VLMs using the aforementioned abstract representations to semantically ground the robot's behavior in the environment. To benchmark and fine-tune VLMs for robotic motion understanding, we introduce the \DatasetName dataset containing 653 human and 369 robot demonstrations across 13 task categories. \MethodName assesses the success of robot motion given the image observation of the trajectory, task instruction, and motion description. Our model significantly outperforms state-of-the-art VLMs by at least twice in precision and $56.1\%$ in recall, generalizing across unseen motions, tasks, and environments. Finally, we demonstrate practical applications of \MethodName in refining and terminating robot planning, and ranking trajectories on how they align with task and motion descriptions.\\
Project page: \href{https://motif-1k.github.io/}{https://motif-1k.github.io/}\vspace{-0.2cm}
\end{abstract}


\begin{figure*}[t!]{
\centering
\begin{center}
\includegraphics[width=1.0\linewidth]{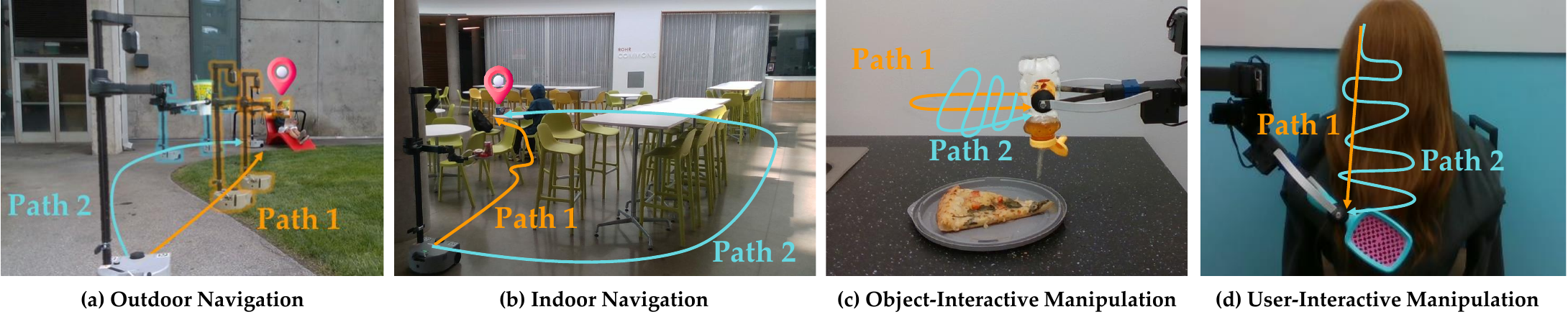}
\end{center}
\centering}
\vspace{-0.45cm}
\caption{\textbf{Different robotic motions for various tasks.} For each task, we visualize two different motions (path 1 and 2) from real robot demonstrations, where the trajectories share the same initial and final states. Most existing success detectors ignore intermediate states, thereby cannot distinguish them.
\vspace{-0.5cm}
} \label{fig:different-motions}
\end{figure*}

\section{Introduction}
Measuring success in robotics has focused primarily on \textit{what} robots should do, not \textit{how} they should do it. Concretely, \textit{what} is determined by the \textit{final state} of an object, robot, or end-effector~\cite{yenamandra2023homerobot, du2023vlm-success-detector}. However, not all trajectories that achieve the same final state are \textit{equally successful}. 
When transporting a fragile object, a path through safer terrain could be considered \textit{more successful} than a shorter yet riskier route (\cref{fig:different-motions} \orangetext{a}). Similarly, in the presence of humans a robot's actions when navigating, holding objects, or brushing human hair (\cref{fig:different-motions} \orangetext{b-d}) can cause surprise, discomfort, or pain, making such motions \textit{less successful}. 

Success detectors play an important role in robot learning since they evaluate whether or not a robot has completed a task. However, most overlook the importance of \quotes{\textit{how}} the task is accomplished, focusing on the initial and final states of the trajectory~\cite{du2023vlm-success-detector, guan2024tasksuccess}.
This simplification fails to account for tasks that fundamentally require evaluating the entire trajectory to assess success.
As we incorporate robots into everyday scenarios, the manner in which they complete tasks will become increasingly important given the context of a scene and its semantic grounding (\textit{e.g.,} moving safely to avoid collisions with chairs). 
Therefore, a more holistic approach to success detection is needed that considers the entire trajectory, the task, and how the agent should move to complete said task.

While modern vision-language models (VLMs) have recently been used as promising tools for success detection~\cite{du2023vlm-success-detector, guan2024tasksuccess}, they are unable to capture complex notions of how a task is completed.
This challenge stems from the fact that VLMs are generally designed to reason over single images, while success detection in robotics is inherently sequential. To bridge this gap, we explore the choice of abstract motion representations, such as visualizing trajectories or aggregating multiple frames in a single image, to better leverage the capabilities of VLMs. In this paper, we propose a trajectory based visual motion representation which overlays a robot's trajectory on the final frame, capturing both the path shape and its semantic connections to the environment. This approach leverages the world knowledge encoded in VLMs and refines it to assess robotic behaviors more effectively.

We propose \textbf{motion instruction fine-tuning}, a method that fine-tunes pre-trained VLMs to equip the capability to distinguish nuanced robotic motions with different shapes and semantic groundings. Given a single frame with a robot's trajectory overlaid and its corresponding task and motion specifications, we query VLMs for success detection. Specifically, our model outputs a binary value indicating whether the motion is \textit{correct} (1) or \textit{incorrect} (0).

We collect the \textbf{\DatasetName} dataset to fine-tune VLMs, due to limited public robot data with diverse semantically grounded motions. We find that co-training mostly on human data with limited robot data enables the model to transfer its knowledge to robotic motion understanding effectively. Based on this insight, \textbf{\DatasetName} contains a variety of motions with 653 human and 369 robot demonstrations across 13 task categories, offering extensive coverage of both the \textit{what} and the nuanced \textit{how} of motion, complete with detailed annotations. It identifies common types of motions featuring varying degrees of semantic grounding, such as the robot's relationship with objects or humans in the environment. The dataset also captures diverse path shapes, in terms of directionality, concavity, and oscillation. For instance, paths in \cref{fig:different-motions} \orangetext{(a)} differ in terms of directionality and spatial relationship with the environment, where it might be undesirable for a robot to pass over the grass. Paths in \cref{fig:different-motions} \orangetext{(d)} demonstrate how straight and curly hairs might require different brushing techniques regarding horizontal oscillations. Notably, \DatasetName includes subtle motions that are often indistinguishable solely by their start and end states (\cref{fig:dataset-all-motions}). 

\textbf{\MethodName}, a motion discriminator developed by fine-tuning on \DatasetName, shows further improved success detection on nuanced robot motions.
We evaluate \MethodName on the validation and test splits of \DatasetName and demonstrate generalization to unseen motions, tasks, and environments. We significantly outperform state-of-the-art VLMs (\textit{e.g.} GPT-4o, GPT-4V and Gemini-1.5 Pro), with at least twice and $56.1\%$ higher precision and recall, respectively. %
Additionally, we demonstrate three applications of using \MethodName: (1) using it as a feedback to finish robot planning, (2) using it as a feedback to refine planning without additional human effort, and (3) using the model output to rank trajectories. 
In summary, our main contributions include:

\begin{itemize}[leftmargin=10pt,noitemsep]
\item \textbf{Motion Representation}: We propose overlaying a robot's trajectory on the final frame to capture the path shape and semantic context, enhancing VLMs' ability to assess success on nuanced robotic behaviors.
\item \textbf{\DatasetName Dataset}: We introduce a dataset with 653 human and 369 robot demonstrations across 13 tasks, providing extensive coverage of various motion types for effective VLM fine-tuning.
\item We develop \textbf{\MethodName}, a motion discriminator fine-tuned on \DatasetName, achieving significant improvements in success detection precision and recall over state-of-the-art VLMs.
\end{itemize}


\section{Related Work}
\label{sec:related-work}
With the recent development of large language models (LLMs) and VLMs, foundation models have been used to understand the environment and critique agent behaviors. Additionally, the increasing use of visual observations in robotics has brought attention to motion-centric visual representations.

\textbf{Success Detection Using Foundation Models.}
\cite{ma2023eureka, kwon2023reward, hu2023language, yu2023language, hwang2023promptable} use LLMs to generate reward functions that evaluate agent behaviors. However, as LLMs cannot understand how a robot's actions are visually grounded in a scene, such methods require everything including the state of the environment to be translated through language. VLMs have been used to understand and evaluate agent behaviors given visual and language information. \cite{lee2024prometheus-vision} use a VLM critic for general vision-language tasks,
and closed API-based models have been used as behavior critics in robotics, even if inaccurate \cite{wang2024rl-vlm-f, guan2024tasksuccess}.
Prior work~\cite{roboclip, wang2024rl-vlm-f, rocamonde2023vision, baumli2023vision} have also used VLMs as zero-shot reward models for training downstream policies. 
\cite{du2023vlm-success-detector} trains a VLM success detector for evaluating what was achieved from the robot, but does not consider \textit{``how''} the agent solves the task. 
\cite{ma2022vip, sermanet2016unsupervised} use representation learning to train reward models as behavior critics.
Other studies~\cite{gao2023physically, chen2024spatialvlm, michal2024robotic} train VLMs to be physically or spatially grounded, but focus on static environment understanding, not dynamic motions. \cite{belkhale2024rth} uses robot joint states to generate motion descriptions in language, but their limited vocabulary is restricted to short horizon, non-grounded motions (e.g., move arm up, rotate arm right). In contrast, we fine-tune VLMs to understand and evaluate grounded motions (e.g., make a detour to the left of the table) and consider more complicated, long-horizon, object or human interactive motions. Our work extends these foundations by specifically using VLMs as success detectors in the context of grounded motion in various tasks.

\textbf{Motion-Centric Visual Representations.}
A growing interest in motion-centric visual representations in robotics has led to recent work in egocentric trajectory representation~\cite{krantz2021waypoint, gu2023rt-traj} and point tracking using optical flow~\cite{doersch2022tap, doersch2023tapir} or learned predictors~\cite{bharadhwaj2024track2act, vecerik2023robotap, wen2023anypoint}. Prior works summarize trajectory or actionable choices in a single image frame by visualizing waypoints~\cite{krantz2021waypoint}, keypoints in the environment~\cite{yuan2024robopoint, liu2024moka}, or desirable paths~\cite{gu2023rt-traj, nasiriany2024pivot, niu2024llarva}; however, the focus of these works is mostly on conditioning the policy on these representations to improve policy learning rather than evaluation of nuanced motions and behaviors.
While Egocentric views are easy to visualize robot trajectories in the camera frame, they often fail to provide comprehensive environmental context and understanding of the robot's movement in the global coordinate. In this paper, we mainly employ visual representations of trajectories overlaid on single images from an exocentric view, which can offer a broader perspective, crucial for analyzing the robot's interactions and movements comprehensively. 

\begin{figure*}[t!]{\centering
    \subfloat[\centering Single Keypoint]{ 
            \includegraphics[width=0.17\linewidth]{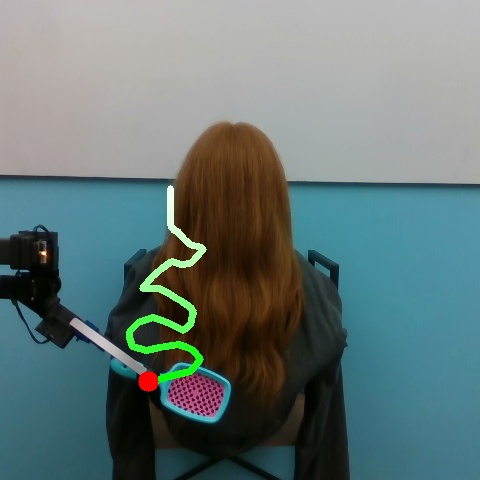}
            }\centering
    \subfloat[\centering Optical Flow]{
            \includegraphics[width=0.17\linewidth]{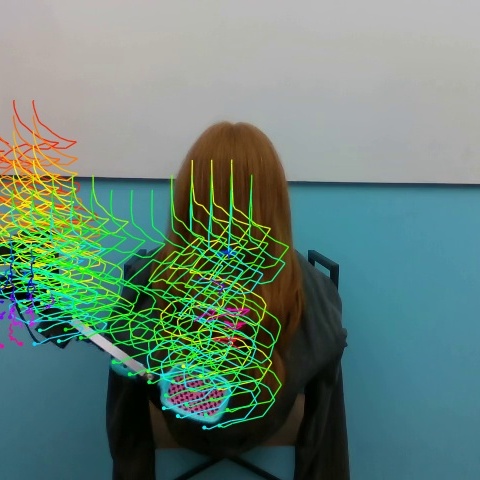}
            }\centering
    \subfloat[\centering 2-frame Storyboard]{ 
            \includegraphics[width=0.355\linewidth]{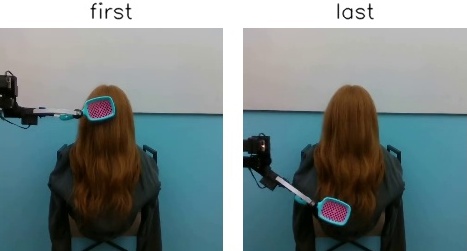}
            }\centering
    \subfloat[\centering 4-frame]{
            \includegraphics[width=0.167\linewidth]{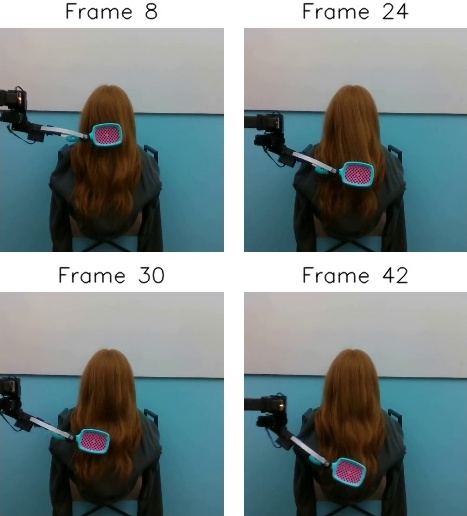}
            }\centering
    \caption{\small{\textbf{Visual Motion Representations} 
    We explore three visual motion representations: (a) single keypoint tracking, (b) optical flow, and (c-d) multi-frame storyboard. For single keypoint tracking, temporal changes are shown with color gradient from white to green, ending with a red circle. For optical flow, we visualize the flow of all keypoints with rainbow colors. We sample $N$ keyframes for $N$-frame storyboard.}
    \vspace{-14pt}
    }
    \label{fig:motion-representations}
}
\end{figure*}

\section{Motion Instruction Fine-tuning (\MethodName)}
\label{sec:method}

In this section we first broaden the definition of success by including motion as a core component. We then discuss why pre-trained models are insufficient for motion-based success detection in robotics. Finally, we introduce \MethodName for fine-tuning VLMs to be motion-aware success detectors.

\vspace{-5pt}
\subsection{Problem Statement}
\label{section3.1-problem-statement}

Success detection has been an integral part of recent robotics literature. Typically, success detection is defined as a binary function of the final state conditioned on the task, $y=f(o_T|\text{task}) \in \{0 , 1\}$~\cite{du2023vlm-success-detector}, where $y$ is the binary success label and $o_T$ is the image observation of the agent and the environment at the final time step $T$. This restrictive assumption prevents success detectors from criticizing \textit{how} a task is completed. For many tasks like collision-aware navigation, 
the final state does not provide sufficient information to capture the robot's interaction with objects or humans in the environment.
Such tasks are often described by both their objective (i.e. bring me lemonade) and their execution (i.e. avoid going over the grass). In these scenarios, success cannot be determined by just the final state. Perhaps the simplest approach is modeling the entire trajectory, $y=f(o_1, \dots, o_T|\text{objective}, \text{execution})$; however, this is computationally costly and often redundant.
We instead propose using an abstract visualization of the trajectory $\tau = (o_1, \dots, o_T)$. Notably, $I(\tau)$ outputs a 2D representation of the full trajectory overlaid on the last image $o_T$.

Suppose a task instruction $T$ and motion description $M$ corresponds to the robot's trajectory. Our goal is to discern how aligned different types of motions are given a task specification to be able to assess success based on \emph{how} the task is done. Doing so requires assessing motion or path shape and, if the task requires object or human interaction,  semantically grounding the robot's motion in the environment.
Thus, we define the set of motions based on two criteria: \textit{path shape} and \textit{semantic grounding} (see \cref{fig:motions-and-prompt}). First, motions are distinguished based on properties of \textit{path shapes} such as directions of translations, rotations, oscillations, repeated motions, and the convexity of curves. Second, \textit{semantic grounding} in the environment involves understanding the context of the scene and the robot's interaction with the environment. For instance, moving over or making a detour towards an object implies the robot is aware of that instance. Similarly, we consider the relative distance and orientation of the agent with respect to the key objects in the scene (see \cref{tab:task-list}). In the following sections, we develop a VLM that acts as a success detector $y=f(o_1, \dots, o_T|T, M)$.
\vspace{-4pt}

\subsection{Visual Motion Representations}

\textbf{Can foundation models be used for success detection?}
While VLMs have demonstrated a strong understanding of physical and causal commonsense reasoning~\cite{AAAI:Bisk2020,ACL:Zellers2019} and semantic grounding~\cite{chen2024spatialvlm, gao2023physically, kwon2023toward}, 
they typically work with static images as inputs, and cannot reason about sequential inputs necessary for dynamic tasks in robotics. Understanding motion requires not only isolating the most meaningful aspects of the scene but also identifying which changes that occurred due to the robot's motion are semantically relevant to the task. Without the ability to understand the changes over time, VLMs may struggle to detect success over robotic motions.

While the ideal model would simply extract semantic content from videos, the strongest current VLMs instead rely on detecting differences between multiple frames, often in the form of storyboards (see \cref{fig:motion-representations} \orangetext{(c-d)} and the example of GPT-4o in \cref{app-sec:experiments}). Though straightforward, this approach often struggles as storyboards lead to lower resolution images. Instead, we borrow insights from prior work~\cite{gu2023rt-traj, nasiriany2024pivot} that show VLMs can effectively leverage diagrams or abstract representations on top of image observations.

\textbf{Representing a robot's motion in a single image.}
 Due to the limitations of storyboards, we explore what representations effectively capture a robot's motion in a single frame.
 To construct diagrams of robotic motions, we overlay a robot's trajectory on the image observation as shown in \cref{fig:motion-representations}  \orangetext{(a-b)}.
One solution is to detect $K$ keypoints $\{(x^1_0, y^1_0), \dots, (x^K_0, y^K_0)\}$ in the initial frame $I_0$, and track the movement of each keypoint until the final frame $I_T$ (see \cref{fig:motion-representations} \orangetext{(b)}). Here, $x^k_t$ and $y^k_t$ denote the x and y coordinates of the $k^{th}$ keypoint in a 2D image observation at timestep $t$, respectively. The detected visual traces, i.e., optical flow, represent the apparent motions of the robot and how the environment changes accordingly. While this solution helps a single image representation contain the information of multiple keypoints, the full optical flow with trajectories of all keypoints may obscure a large portion of the background and important objects in the scene. Additionally, visualizing many keypoints often results in indistinguishable or overlapping trajectories, which may create visual clutter and reduce the clarity of the motion. 

Therefore, the proposed method, \MethodName, visualizes the trajectory of the most representative keypoint (see \cref{fig:motion-representations} \orangetext{(a)}), as a simplified yet more interpretable visual motion representation. We call the keypoint as point of interest, where a point of interest is typically chosen as a point on the end effector's surface by human annotators. 
Compared to prior work~\cite{gu2023rt-traj} that visualizes the center of mass of the end effector, our approach ensures that the selected keypoint is visually recognizable and not occluded. Details on labeling the point of interest and visualizing its trajectory are provided in \cref{sec:dataset}.

\vspace{-1pt}
\subsection{Fine-Tuning VLMs}\label{sec:method-fine-tuning-vlms}

\begin{figure}[t!]{
\centering
\begin{center}
\includegraphics[width=0.8\textwidth]{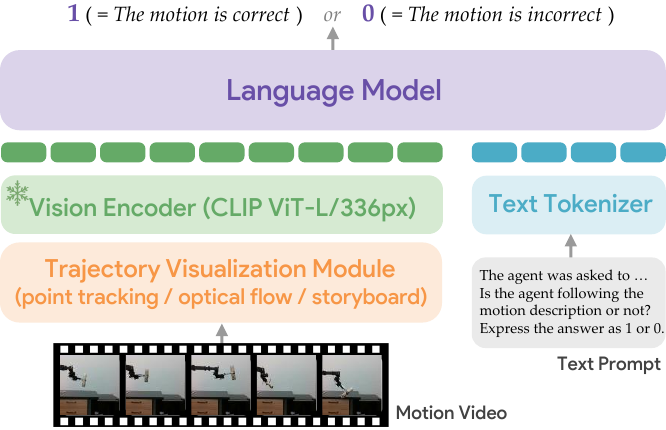}
  \end{center}
  \caption{\textbf{Network Architecture.}  Given a visual motion representation of a robot’s trajectory and its corresponding task and motion specifications, our model outputs a binary value indicating whether the motion is {correct} (1) or {incorrect} (0). 
  \vspace{-15pt}
  } \label{fig:network-architecture}
  }
\end{figure}

While we can directly pass a single image visualizing the robot's trajectory into a model zero-shot, non-fine-tuned models struggle to understand complex robotic motions. \cref{sec:experiments} empirically shows that existing state-of-the-art VLMs often infer undesirable motions as correct (false positives) or successful motions as incorrect (false negatives).
Prior works~\cite{gao2023physically, chen2024spatialvlm} have shown promising results fine-tuning VLMs to improve their grounding capabilities and understanding of a scene. Similarly, we also fine-tune VLMs with the different representations proposed earlier.

\textbf{Model Architecture.}
\cref{fig:network-architecture} shows the network architecture of our model. Given a trajectory representation $I(\tau)$ as an image and a text prompt consisting of a task instruction $T$ and motion description $M$ followed by the question \textit{"Is the agent following the motion description or not? Express the answer as 1 or 0."}, our model outputs a binary success prediction $y$.
The prediction $y$ indicates whether the motion is \textit{correct} (1) or \textit{incorrect} (0) with respect to the task and motion specification $(T, M)$. During fine-tuning, only the language model is updated, while the visual encoder is frozen.

\textbf{Constructing Training Data.}
Effective fine-tuning requires a well-structured dataset that includes both positive and negative samples. Positive samples consist of trajectories paired with their corresponding task instructions and motion descriptions. This encourages the model to associate visual motion representations with their correct task specification.
For the $i^{th}$ trajectory $\tau_i$ in the training set we generate a single image $I_i$ representing the trajectory based on the chosen visual motion representation. To construct positive samples for training, we pair the set of $K$ images with their corresponding task instructions and motion descriptions, forming the set $\mathbf{S}^+_{\text{train}}=\{(I_1, T_1, M_1), \dots, (I_K, T_K, M_K)\}$, where $T_i$ and $M_i$ are the task instruction and motion description for image $I_i$, respectively. To construct negative samples for image $I_i$, we choose the $N_{\text{neg}}$ least similar motion descriptions. The similarity between motion descriptions is calculated using SentenceTransformer~\cite{reimers-2019-sentence-bert} embeddings. The set of negative samples is then constructed as $\mathbf{S}^-_{\text{train}}=\bigcup_{i=1}^{K}\{(I_i, T_i, M^-_{i, 1}), \dots, (I_i, T_i, M^-_{i, N_{\text{neg}}})\}$, where $M^-_{i,j}$ is the $j^{th}$ least similar motion description to $M_i$. We set $N_{\text{neg}}=10$ in our experiments. The full training dataset $D$ is $\mathbf{S}^+_{\text{train}}\cup\mathbf{S}^-_{\text{train}}$.

\textbf{Co-training with Human and Robot Data.}
Based on prior work~\cite{liu2023llava}, fine-tuning VLMs require a huge amount of training data.
Ideally our dataset $D$ would cover all motions that a robot would execute across a wide variety of tasks. Unfortunately, robot demonstrations are time-consuming and often difficult to collect. This makes scaling the size of the dataset hard when we only rely on robot data for fine-tuning VLMs. Moreover, robots may have physical constraints which prevent a model trained with a specific robot embodiment from generalizing to other robots with different dynamics. On the other hand, human demonstrations have a high degree of freedom and are often easier and more intuitive to collect, especially when looking for a diversity of motion. Thus, we opt to train on a mixture of human and robot demonstrations, $D_h$ and $D_r$ respectively. By doing so, we facilitate easy data collection while also ensuring the downstream VLM is more robust to embodiment and motion types.


\section{\DatasetName Dataset}
\label{sec:dataset}
\begin{figure}[h!]{
\centering
  \begin{center}
\includegraphics[width=0.96\textwidth]{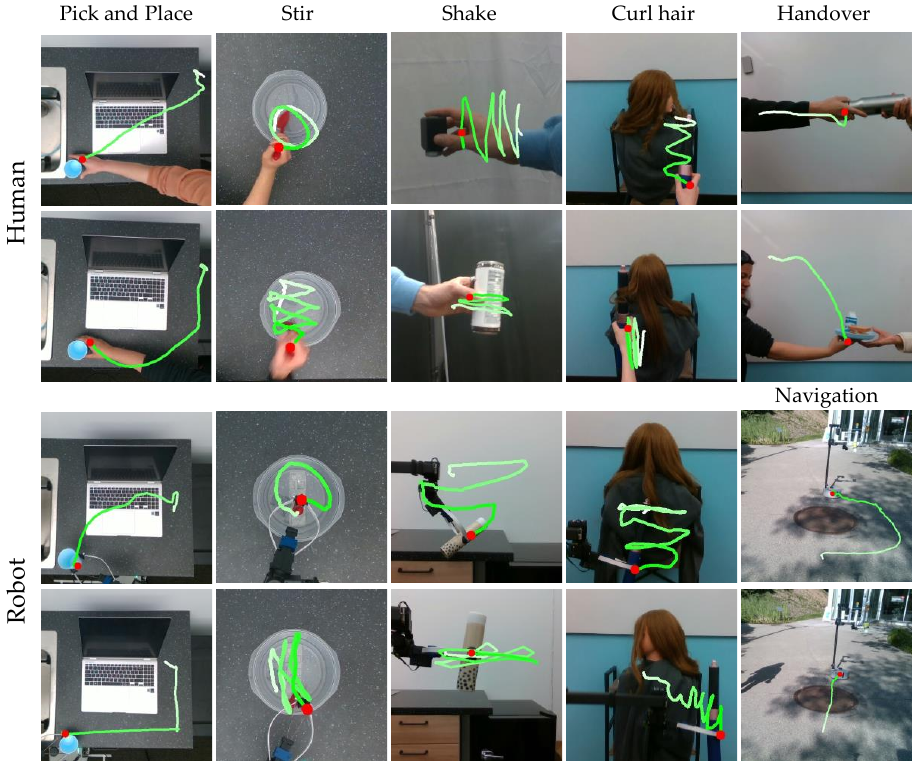}
  \end{center}
  \caption{\small{\textbf{Trajectory Visualizations.} 
  We visualize two different motions for solving the same task with the same embodiment. }
  }
  \label{fig:dataset-all-motions}
}
\end{figure}

To benchmark and improve VLMs for motion understanding, we release the \DatasetName dataset containing 653 human and 369 robot demonstrations across 13 tasks.
As in \cref{fig:different-motions} and \cref{fig:dataset-all-motions}, each task has demonstrations for $2$ to $5$ distinct motions which vary during the intermediate steps of a trajectory.  For instance, a motion's path shape and semantic relationship with nearby objects may be different across demonstrations. This captures the diversity of \textit{how} a task can be achieved, reflecting the nuanced and complex motions present in real-world scenarios. For instance, when shaking a boba drink, one person might use vigorous vertical movements, while another might use careful side-to-side movements to avoid spillage after inserting a straw. Moving a cup near a laptop via the shortest path is preferred if the cup is empty, but a detour is necessary if the cup contains water. Motion diversity is essential in grounded settings where motions need to adapt to varying environmental contexts. \cref{fig:dataset-all-motions} shows examples of diverse motions. By collecting a diverse set of context-dependent motions with different intermediate trajectories, we ensure that our dataset challenges VLMs to consider the full trajectory for success detection.

\begin{figure*}[t!]{\centering
    \includegraphics[width=1\linewidth]{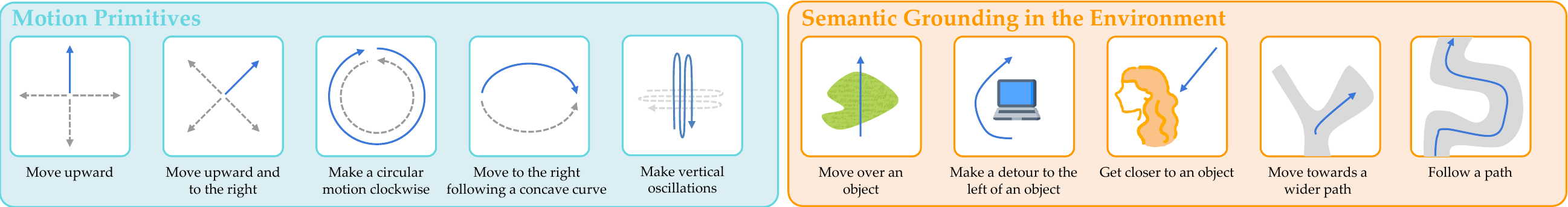}
    \caption{\small{\textbf{Motion Diversity.} 
    10 canonical motions are described with blue arrows and their corresponding descriptions. In the blue box, motion primitives are categorized based on the shape of the ideal path. Gray dashed arrows denote variants of the blue arrow, lying in the same category. The orange box shows motions that involve grounding in the environment, where the relationship between the robot and an instance in the environment is considered.
    }\vspace{-14pt}
    }
    \label{fig:motions-and-prompt}
}
\end{figure*}

\subsection{Collecting Human and Robot Demonstrations}
As addressed in \cref{sec:method-fine-tuning-vlms}, we use a mixture of human and robot demonstrations. In this section, we will explain how we collect human and robot demonstrations. 
\cref{fig:dataset-all-motions} visualizes the final frames of the collected human and robot demonstrations.
Our human demonstrations are collected by six different people to ensure ample variation in motion. For robot data, a single human expert teleoperates a Stretch robot with a Meta Quest 3 VR controller for manipulation tasks and a gamepad for navigation tasks. We record the agent's joint states and image observations with a fixed exocentric RGBD camera for visual consistency. For the pick and place, stir, shake, brush hair, and tidy hair tasks we collect trajectories in two orthogonal camera views to support future 3D motion understanding using multiview image observations. Since we use VLMs that input RGB images, we treat observations from different camera viewpoints as separate trajectories and focus on effectively representing the agent's motion on a 2D image frame. \cref{sec:grounded-motion-annotations} explains how we annotate the trajectories with task instructions and fine-grained motion descriptions.

We preprocess the collected demonstration observations using three motion representation methods (see \cref{fig:motion-representations}):

\begin{itemize}
    \item \textbf{optical flow}~\cite{doersch2022tap, doersch2023tapir}: visualizing the trajectories of all visible keypoints with rainbow colors. For each keypoint, its trajectory is drawn with a single color.
    \item \textbf{single keypoint tracking (\MethodName)}: visualizing a trajectory of a single keypoint. For single point tracking on human data, we use mediapipe~\cite{lugaresi2019mediapipe} and track the center of the hand pose. For robot data, we annotate 2D keypoints to identify point of interests in the initial frame of each episode, which is either a keypoint on the object that is being manipulated or the initial position of the robot's end effector. Then, we choose the keypoint nearest to the point of interest. For both human and robot data, temporal changes are shown with color gradient from white to green, ending with a red circle.
    \item \textbf{$N$-frame storyboard ($N=2,4,9$)}: sampling $N$ keyframes and stacking those frames into a single image. We use K-means clustering on the image embeddings of all frames to sample keyframes that are sufficiently different in latent space. Frame indices are annotated above each frame image.
\end{itemize}

\subsection{Grounded Motion Annotations}\label{sec:grounded-motion-annotations}
In this section, we use ``agent'' to refer to either human or robot demonstrating the task. We use human annotators to label the motion in each video. While automated motion labeling using proprioception has previously been used, it is only applicable to short-horizon motions (<10 frames), whereas our dataset contains long-horizon motions (>300 frames).
Compared to previous datasets which do not capture any information about \textit{how the motions are grounded in the environment or to the user},
we consider motion diversity in two different axes: (1) path shape, and (2) semantic grounding in the environment.

\textbf{Path Shape.} As illustrated in \cref{fig:motions-and-prompt}, we first set a vocabulary of motion primitives in terms of path shape; direction and convexity of translation (e.g., move upward, follow a convex curve), direction of rotation (e.g., make a circular motion clockwise), and oscillatory movement (e.g., move up and down). Motion often consists of multiple motion primitives with different path shapes, such as moving right then downward, or moving right while oscillating vertically. When annotating such motions in language, we first list the primitives in the temporal order and prioritize the dominant ones among those that happen simultaneously.

\textbf{References to Objects and Humans.}
Another important aspect of grounded motions are references to objects or humans in the scene. 
The orange box in ~\cref{fig:motions-and-prompt} illustrates five common examples of motions in terms of semantic grounding. These motions gain specific meaning in relation to objects or humans in the environment.
Based on the task specification regarding an object (e.g., avoid damaging the laptop) or a user (e.g., focus brushing the bottom part of the hair that is tangled), an agent's motion can be distinguished accordingly.
For instance, a motion of shaking salt over food is defined not only by the path shape of the shaking movement but also by the spatial relationship between the agent and the food. We use a set vocabulary (e.g., move over, make a detour, get closer/farther, follow path) to annotate the grounded motions in language.

\begin{figure}[t!]{
\centering
  \begin{center}
\includegraphics[width=\textwidth]{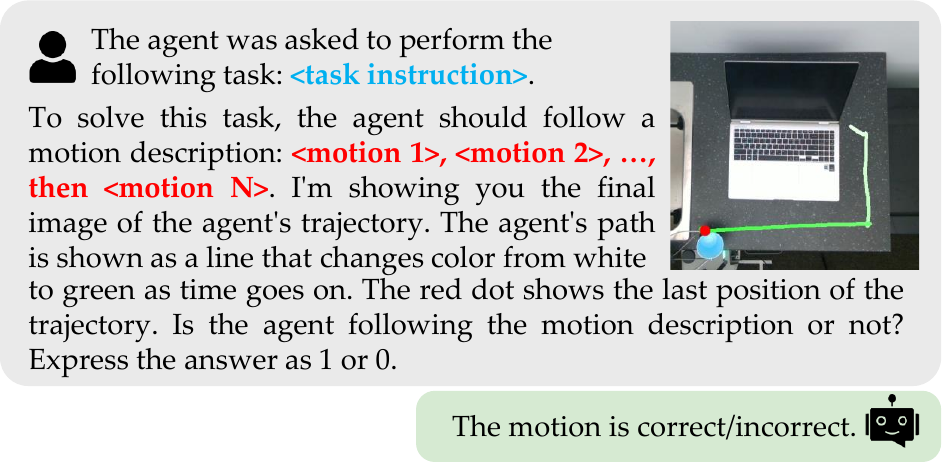}
  \end{center}
  \vspace{-10pt}
  \caption{
  \small{\textbf{Text Prompt Template} shows how a motion composed of multiple primitives is described in the text input of a VLM. }\vspace{-8pt}
  }\label{fig:text-prompt-template}
  }
\end{figure}

The final motion descriptions in \DatasetName are constructed as combinations of descriptions in terms of path shape and references to objects and humans.
With the annotated task instructions and motion descriptions, we construct text prompts that could be used to describe the agent's motion. Using this text prompt and the image representation of the trajectory as inputs, we train VLMs to predict the alignment between the text and image (see \cref{sec:method-fine-tuning-vlms}). For the example in \cref{fig:text-prompt-template}, the task instruction is ``pick up the cup and place it to the lower left of the laptop'' and the motion description could be ``move downward and farther from the laptop, then move to the left''. Here, two motion primitives, ``move downward'' and ``move to the left'', are combined with a grounded motion annotation,``move farther from the laptop''. Given the text prompt and trajectory representation $I(\tau)$ in \cref{fig:text-prompt-template}, our fine-tuned VLM outputs a binary value indicating whether the motion is correct or not.

\section{Experiments}
\label{sec:experiments}
In this section we seek to answer the following questions: 1) How does MotIF compare to start-of-the-art models? 2) How important is robot data in understanding motion? and finally 3) What is the effect of visual motion representation?
We compare our approach with state-of-the-art models and assess the benefits of co-training on human and robot data. We investigate the impact of different visual motion representations. All models are evaluated on the validation and test splits in \DatasetName. 
See \cref{app-sec:experiments} for detailed analyses.

\textbf{Baselines.} We evaluate against GPT-4o, GPT-4V~\cite{achiam2023gpt4}, and Gemini-1.5 Pro~\cite{team2023gemini} as state-of-the-art API-based baselines. We also compare to the best performing pre-trained open LLaVA~\cite{liu2023llava} models with various sizes (7B, 13B, 34B). To evaluate different visual motion representations, we compare our proposed single point tracking to full optical flow, $N$-frame storyboard ($N=2,4,9$), and a single-frame image. All learned baselines are fine-tuned from the LLaVA-v1.5 7B model \cite{liu2023llava, liu2023improvedllava}, which was pre-trained on the Vicuna visual question answering (VQA) dataset~\cite{chiang2023vicuna}. 

\textbf{Training Details.}
We train on all human and 100 robot demonstrations from \DatasetName. For each demonstration, we construct one positive and 10 negative samples of image-text pairs. When fine-tuning the VLMs, we freeze the weights of the pre-trained visual encoder, CLIP ViT-L/14~\cite{radford2021vit} with an input size of 336px by 336px. We fine-tune the projection layer and the language model in the VLM using low-rank adaptation (LoRA) with cross-entropy loss for 30 epochs with a learning rate of $5e^{-5}$ and a batch size of 32. We use a single A100 GPU for fine-tuning. 

\textbf{Evaluation Set and Metrics.}
All VLMs output a binary label indicating whether or not the agent's motion in the image aligns with the given task and motion descriptions. For off-the-shelf models, we convert natural language responses into their corresponding binary labels. We evaluate models on the validation and test split of \DatasetName containing $129$ and $134$ robot demonstrations, respectively. The test split contains a set of unseen trajectories which vary in terms of camera viewpoints, motions, tasks, and environment in comparison to the training and validation split. Similar to the training data, we construct one positive and $10$ negative samples per demonstration.
For each experiment, we evaluate the performance of models using precision {\small$(=\text{TP}/(\text{TP}+\text{FP}))$} and recall {\small$(=\text{TP}/(\text{TP}+\text{FN}))$}\footnote{TP: True Positive, FP: False Positive, TN: True Negative, FN: False Negative}.
These metrics show the reliability and robustness of model outputs. High precision minimizes false positives, and high recall ensures most valid motions are identified.

\begin{figure*}[t!]{\centering
    \subfloat[\centering Comparison with SOTA off-the-shelf VLMs]{
            \includegraphics[width=0.55\linewidth]{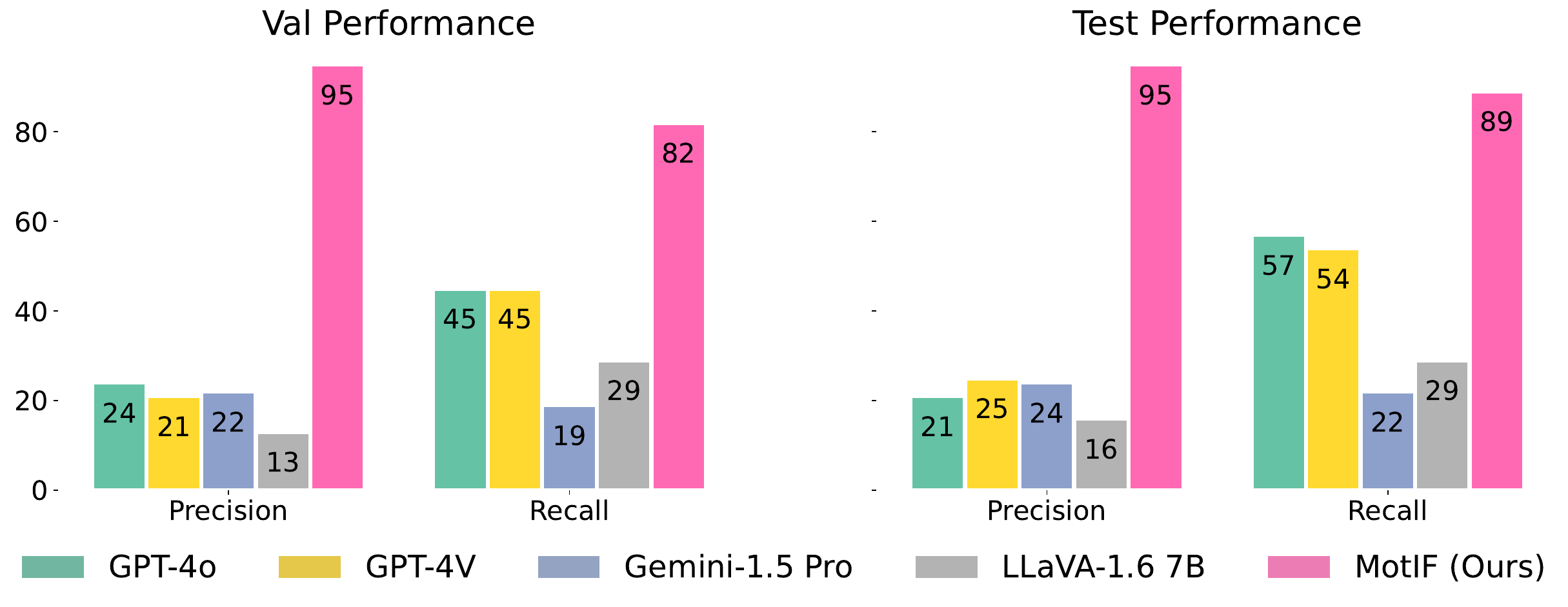}
            }\centering
    \subfloat[\centering Motion Representations]{
            \includegraphics[width=0.4\linewidth]{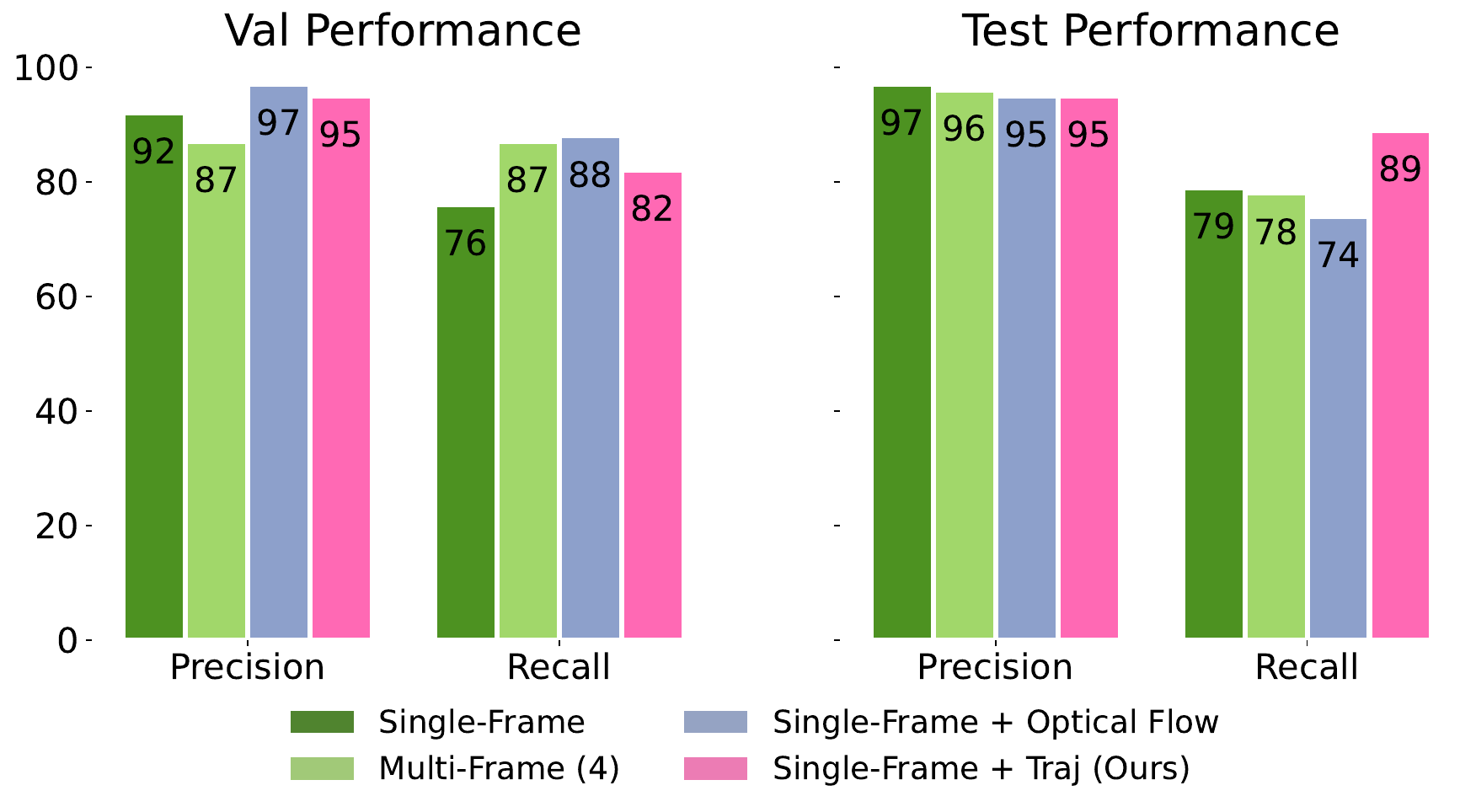}
            }\centering
    \vspace{-5pt}
    \caption{
    \small{\textbf{Performance on \DatasetName.} (a) shows that our models outperforms state-of-the-art (SOTA) off-the-shelf models in validation and test splits.
     (b) We explore motion representations in terms of single-frame vs. multi-frame and the effectiveness of trajectory drawing. Single-frame with trajectory drawing demonstrates the highest recall in the test split, while other motion representations falter. Our approach identifies valid motions effectively and generalizes better than baselines. Among the three multi-frame representations ($N=2,4,9$), we report the performance of the model that performs the best in the test split ($N=4$).
    }\vspace{-10pt}
    }
    \label{fig:performance-sota-and-cotraining}
}
\end{figure*}

\textbf{How does \MethodName compare to state-of-the-art VLMs?}
Our model is developed by fine-tuning a 7B LLaVA model with the \DatasetName dataset. While LLaVA is pre-trained on general VQA tasks answering questions given a static image, we fine-tune the model using a motion representation that overlays a single keypoint trajectory on the final image.
As shown in \cref{fig:performance-sota-and-cotraining} \orangetext{(a)}, \MethodName outperforms GPT-4o, GPT-4V, Gemini-1.5 Pro, and LLaVA in both precision and recall in the test split, by at least $280.0\%$ and $56.1\%$, respectively. We include results in the Appendix that show our model robustly works in data out of training domain, such as unseen tasks and environments.

\begin{figure}[ht!]{
\centering
  \begin{center}
\includegraphics[width=1.0\textwidth]{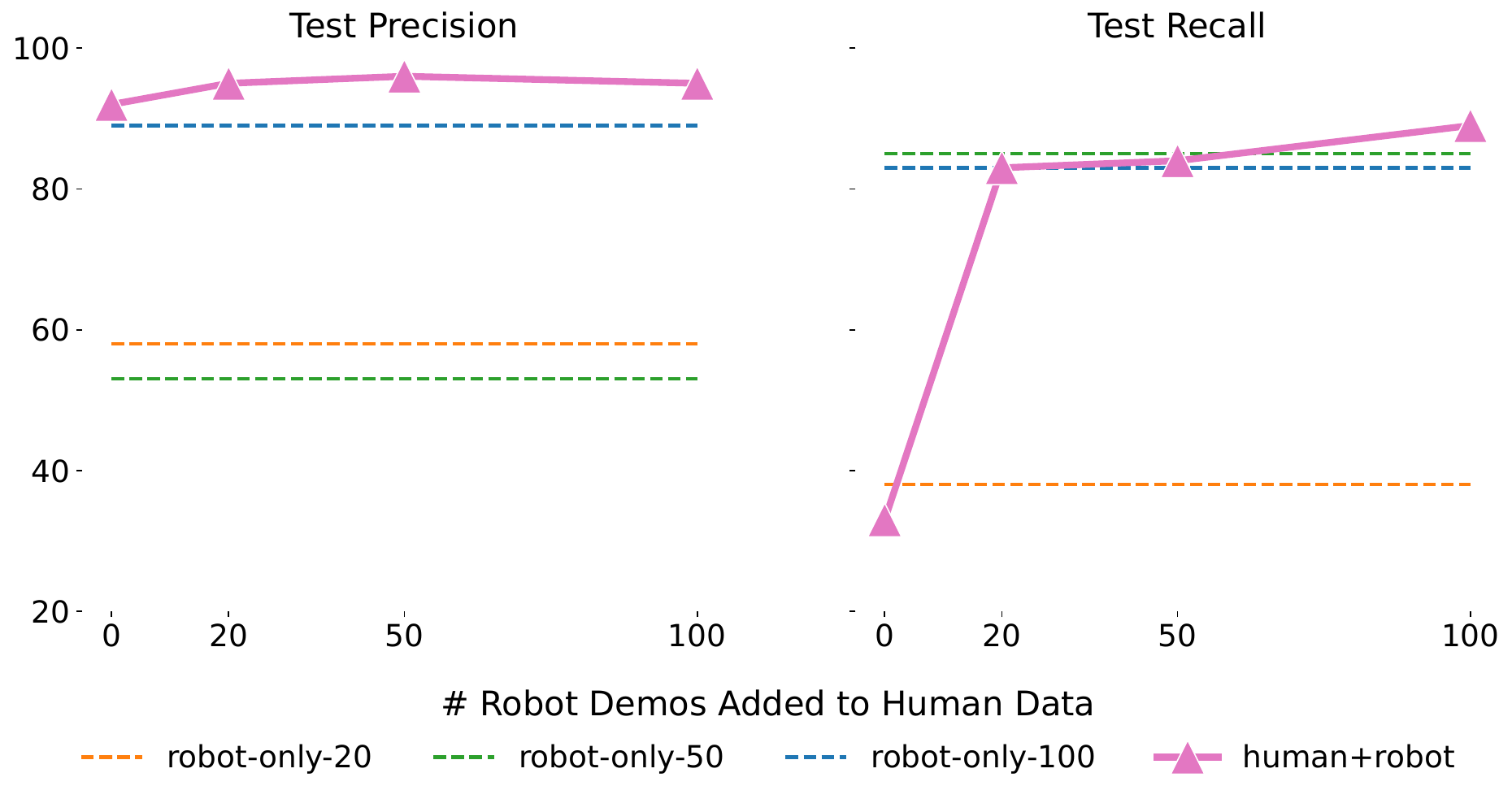}
  \end{center}
  \vspace{5pt}
  \caption{\small{\textbf{Co-training.} Performance improves with more robot demonstrations. Dashed lines indicate performance with robot-only data. Performance of the model trained only with human data is shown as human+robot with \#robot=0 (the leftmost pink triangle).}\vspace{-10pt}}
  \label{fig:co-training-graph}
  }
\end{figure}

\textbf{How does robot data impact performance?}
\cref{fig:co-training-graph} shows how our model results in positive transfer from human to robot data, by co-training on full human data with 653 trajectories and a small subset of robot data. The plots show how precision and recall improve when adding 0 to 100 robot demos. Simply adding 20 robot demos improved performance significantly in recall by $151.5\%$ and slightly in precision by $3.3\%$. With 50 and 100 robot demos, performance improves in all evaluation metrics. 
Interestingly, co-training on human data and 20 robot demos outperforms training solely on 20 robot demos by $63.8\%$ and training on 100 robot demos by $6.7\%$. This implies that human data can be used to learn representations of grounded motions which can transfer to robots, despite the large embodiment gap. However, some robot data is still necessary; training only on human data shows very poor performance ($33\%$ recall) which is worse than random guessing.
This might be because fine-tuning exclusively on human data specializes the model towards the human domain, which may cause performance decrease on robot data. 
Adding even a small amount of robot data significantly improves performance by encouraging the model to learn more generalized, embodiment-agnostic representations, demonstrating the crucial role of robot data in achieving robust performance.

\textbf{How does motion representation affect performance?}
\cref{fig:performance-sota-and-cotraining} \orangetext{(b)} compares different visual motion representations trained on all human data and 100 robot trajectories used in the co-training experiment. Using the last frame without any motion representation obtains recall scores lower than $80\%$ in validation and test splits. In contrast, using \MethodName (single frame + Traj) improves recall by $7.9\%$ and $12.7\%$ on the validation and test split, respectively. Using full optical flow is effective in the validation split, but it generalizes poorly to test split, performing worse than \MethodName. Perhaps the simplest baseline, inputting multiple frame storyboard, also fails to outperform \MethodName. Performance degradation with multi-frame and optical flow is likely due to reduced image quality from either reduced image resolution or visual clutter from optical flow that covers relevant information.

\begin{figure*}[t!]{\centering
    \subfloat[\centering Refining and Terminating Robot Planning]{
            \includegraphics[width=0.53\linewidth]{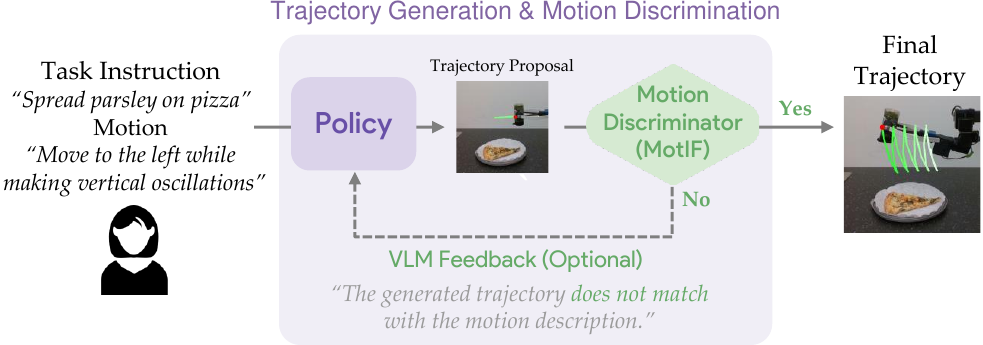}
            \vphantom{\includegraphics[height=0.08\textheight,valign=c]{images/motif_robot_planning.pdf}}
            }\centering
    \hspace{10pt}
    \subfloat[\centering Ranking Trajectories]{
            \includegraphics[width=0.38\linewidth]{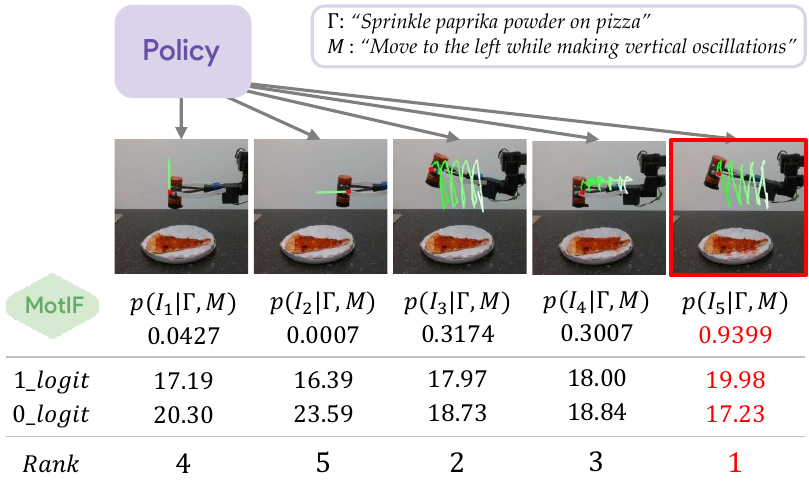}
            }\centering
    \caption{\textbf{Examples of using \MethodName in robot control.} (a) \MethodName can close the loop of any existing open-loop controlled system by automatically determining success on executing proper motion and giving this as a feedback to the system. We use an LLM as the policy that generates the sequence of the robot joint states. (b) We can use \MethodName to rank trajectories, which is a useful application for sampling based planners. $p(I_k|\Gamma, M)$ denotes how likely the motion in the $k^{th}$ image corresponds to the given task instruction $\Gamma$ and motion description $M$. \vspace{-10pt}
    }
    \label{fig:implications-motif}
}
\end{figure*}

\section{Discussion}
\label{sec:conclusion}
\noindent \textbf{Summary.} 
Task specification in robotics often goes beyond simply stating \emph{what} the objective is, and additionally consists of \emph{how} a task should be done. As a step in this direction, we introduce a dataset, \DatasetName, alongside a unique representation method and training technique, \MethodName. Our findings demonstrate that our system can effectively provide assessments of nuanced robotic actions. 
The \DatasetName dataset consisting of human and robot trajectories, captures the diverse ways in which tasks can be executed. We propose \MethodName that uses this data to fine-tune open-source VLMs to detect a more nuanced notion of success. Our results demonstrate that \MethodName can effectively assess success over these nuanced motions by leveraging a simple visual motion representation; overlaying the robot's trajectory on the image observation.

\MethodName can determine the success of trajectories generated by any policy evaluating if the motions align with the task instruction and motion description. This can provide a signal for when to terminate an episode or to further refine the policy. We include examples of the following uses of \MethodName on LLM generated policies~\cite{liang2023codeaspolicies}: \MethodName as a success detector (\cref{fig:overview}), providing correction feedback to the policy, terminating or adapting robot policies (\cref{fig:implications-motif} \orangetext{(a)}), and ranking trajectories (\cref{fig:implications-motif} \orangetext{(b)}).

\textbf{Limitations and Future Directions.}
One limitation of \MethodName is the dependency on 2D visual motion representations, which might not capture all aspects of complex 3D motions. 
Future work could incorporate RGB-D and multi-view images, or 3D visual representations overlaid on the image for a more comprehensive understanding of 3D space using our dataset.
Another direction is to use \MethodName as a reward signal for reinforcement learning. \MethodName can act as a motion discriminator when training RL policies that can potentially more accurately reflect user preferences and contextual appropriateness. 
Future work might create a VLM that outputs natural language responses, which could be used to not only evaluate binary success but also perform reasoning on robotic motions.


{   
    \small
    \def\baselinestretch{0.98}
    \bibliographystyle{unsrt}
    \bibliography{references}
}

\clearpage
\begin{center}
    \small{APPENDIX}
\end{center}

\crefalias{figure}{appendixfigure}%
\renewcommand\figurename{App. Fig.}%
\crefalias{table}{appendixtable}%
\renewcommand\tablename{App. Table}%

\setcounter{page}{1}
\setcounter{section}{0}
\renewcommand\thesection{\Alph{section}} 
\renewcommand\thesubsection{\thesection.\arabic{subsection}} 
\newcommand{\appsubsection}[1]{%
  \refstepcounter{subsection} 
  \noindent\textbf{\thesection.\arabic{subsection}~#1} 
  \par
}

We provide additional details and analyses of \MethodName and \DatasetName in the appendix. \cref{app-sec:dataset-details} provides details for how we construct and annotate the dataset. \cref{app-sec:method-details} provides implementation details of the proposed motion discriminator, \MethodName. \cref{app-sec:experiments} provides additional experiment results and detailed analyses on the results in the paper. \cref{app-sec:implications-motif} provides the implications of \MethodName in real robot planning, with detailed examples.

\section{\DatasetName Dataset Details}
\label{app-sec:dataset-details}
\begin{table*}[ht!]
\setlength{\aboverulesep}{0pt}
\setlength{\belowrulesep}{0pt}
\setlength{\extrarowheight}{.1ex}
\resizebox{\textwidth}{!}{
\begin{tabular}{lllcc}
\toprule
\multicolumn{1}{c}{\multirow{2}{*}{\textbf{Category}}} & \multicolumn{1}{c}{\multirow{2}{*}{\textbf{Task}}} & \multicolumn{1}{c}{\multirow{2}{*}{\textbf{Motion Description Examples}}} & \multicolumn{2}{c}{\textbf{Demonstrations}} \\
\multicolumn{1}{c}{} & \multicolumn{1}{c}{} & \multicolumn{1}{c}{} & \multicolumn{1}{c}{\textbf{Human}} & \multicolumn{1}{c}{\textbf{Robot}} \\
\midrule
\multirow{2}{*}{\begin{tabular}[c]{@{}c@{}}Non-Interactive\end{tabular}} & Outdoor Navigation & \begin{tabular}[c]{@{}l@{}}move in the shortest path\\ make a detour to the left and follow the walkway, avoiding moving over the grass\end{tabular} & - & \checkmark \\ \cmidrule{2-5}
 & Indoor Navigation & \begin{tabular}[c]{@{}l@{}}move in the shortest path\\ make a detour to the right of the long table, avoiding collision with chairs\end{tabular} & - & \checkmark \\ \cmidrule{2-5}
 & Draw Path & \begin{tabular}[c]{@{}l@{}}make a triangular motion clockwise\\ move upward and to the right\end{tabular} & - & \checkmark \\ 
\midrule
\multirow{8}{*}{\begin{tabular}[c]{@{}c@{}}Object-Interactive\end{tabular}} & Shake & \begin{tabular}[c]{@{}l@{}}move up and down 4 times\\ completely flip the object to the right and flip it back to its initial state\end{tabular} & \checkmark & \checkmark \\  \cmidrule{2-5}
 & Pick and place  & \begin{tabular}[c]{@{}l@{}}move downward and to the left\\ move downward while getting farther from \textless{}obstacle\textgreater, then move to the left\end{tabular} & \checkmark & \checkmark \\  \cmidrule{2-5}
 & Stir & \begin{tabular}[c]{@{}l@{}}make 2 circular motions counter-clockwise\\ move upward, then move downward while making diagonal oscillations\end{tabular} & \checkmark & \checkmark \\  \cmidrule{2-5}
 & Wipe & \begin{tabular}[c]{@{}l@{}}move to the right and move to the left, repeating this sequence 2 times\\ move to the right, making diagonal oscillations\end{tabular} & \checkmark & \checkmark \\  \cmidrule{2-5}
 & Open/Close the cabinet & \begin{tabular}[c]{@{}l@{}}move to the right\\ move upward and to the left\end{tabular} & \checkmark & - \\  \cmidrule{2-5}
 & Spread Condiment & \begin{tabular}[c]{@{}l@{}}move to the left and to the right\\ move to the left while making back and forth oscillations\end{tabular} & - & \checkmark \\
\midrule
\multirow{8}{*}{\begin{tabular}[c]{@{}c@{}}User-Interactive\end{tabular}} & Handover & \begin{tabular}[c]{@{}l@{}}move upward and to the left\\ move downward and to the right following a concave curve\end{tabular} & \checkmark & - \\  \cmidrule{2-5}
 & Brush hair & \begin{tabular}[c]{@{}l@{}}move downward while making horizontal oscillations\\ make 5 strokes downward, increasing the starting height of each stroke\end{tabular} & \checkmark & \checkmark \\  \cmidrule{2-5}
 & Tidy hair & \begin{tabular}[c]{@{}l@{}}move downward and to the right following a convex curve\\ make a circular motion clockwise, move upward, then move downward and to the right\end{tabular} & \checkmark & \checkmark \\  \cmidrule{2-5}
 & Style hair & \begin{tabular}[c]{@{}l@{}}move to the right shortly, then move to the left following a concave curve\\ make a circular motion clockwise, gradually increasing the radius of the circle\end{tabular} & \checkmark & \checkmark \\
\bottomrule
\end{tabular}}
\caption{\textbf{List of Tasks and Motion Descriptions.} The collected dataset contains 653 human and 369 robot demonstrations across 13 task categories, 36 task instructions, and 239 motion descriptions. Checkmarks denote which agent (human/robot) demonstrations exist for each task. The table provides two motion description examples for each task.}\label{tab:task-list}
\end{table*}

\appsubsection{Grounded Motion Annotations}
\textbf{Path Shape.} For translation motions, we consider the direction of the movement such as \textit{move upward} or \textit{move upward and to the right}. As motions could be forming a curve or part of a shape, we also consider the convexity of the path or compare the path to figures (e.g., circle, triangle, and square) that resemble the shape being traced. For cyclic motions, the direction such as clockwise/counter-clockwise are considered. Many trajectories may be a composition of simpler motions. Side-to-side motions could be described as \textit{move to the right and to the left N times} or vice versa. When one motion component dominates the overall agent behavior among multiple motion components, we describe the dominating motion first and include the remaining details as modifiers. For instance, a motion can be a combination of downward translation and side-to-side motion, thereby resulting in \textit{move downward, making horizontal oscillations}.

\textbf{Grounding in the Environment.} 
Building on top of VLMs' semantic reasoning capabilities by leveraging their world knowledge, we address how robotic motions can be grounded in the environment. The most basic grounded motion is moving over or avoiding an instance in the environment. As shown in the second leftmost motion in the orange box of ~\cref{fig:motions-and-prompt}, to avoid an obstacle or a hazardous region, the agent can make a detour to the left or right of the corresponding instance. Another example could be getting closer to or farther from an instance, even when the instance is not essentially a target. This could happen when people want legibility or predictability of the agent's motion. The instance could also be a path as shown in the two rightmost motions, such as walkways, drawings on a whiteboard, or spills on a table.

\section{Implementation and Experiment Details}
\label{app-sec:method-details}

\appsubsection{Image Preprocessing}
To preprocess images, we first crop all images to be square. Next, depending on the visual motion representation, the images are sampled and aggregated into a storyboard, or have trajectories of single or multiple keypoints overlaid.

\textbf{Multi-frame Storyboard.}
For constructing multi-frame storyboards, we investigate $N=2,4,9$ numbers of keyframes, where the keyframes are sampled from K-means clustering of $N$ clusters based on the ResNet18 embeddings of the images at each timestep of a given visual trajectory.

\textbf{Optical flow.}
To extract optical flow from a robot's motion video, we use the BootsTAPIR~\cite{doersch2024bootstap, doersch2023tapir} algorithm. We run random sample consensus (RANSAC) with parameters in \cref{tab:optical-flow-parameters}.

\begin{table*}[ht!]
\setlength{\aboverulesep}{0pt}
\setlength{\belowrulesep}{0pt}
\setlength{\extrarowheight}{.1ex}
\resizebox{0.4\textwidth}{!}{
\begin{tabular}{ll}
\toprule
\textbf{Parameter} & \textbf{Value} \\
\midrule
Inlier Point Threshold & 0.01 \\
Minimum Fraction of Inlier Points & 0.95 \\
\# Refinement Passes & 2 \\
Foreground Inlier Threshold & 0.01 \\
Minimum Fraction of Foreground Inlier Points & 0.60 \\
\bottomrule
\end{tabular}}
\caption{\textbf{Optical Flow Parameters.}}\label{tab:optical-flow-parameters}
\end{table*}

\textbf{Evaluation Details.}
For interpretability, the scale of the y-axis in \cref{fig:performance-sota-and-cotraining} and \cref{fig:co-training-graph} are 100 times larger than the actual scale of evaluation metrics; precision and recall.

\section{Additional Experiments and Analyses}
\label{app-sec:experiments}
\begin{table*}[t!]
\setlength{\aboverulesep}{0pt}
\setlength{\belowrulesep}{0pt}
\setlength{\extrarowheight}{.5ex}
\resizebox{\textwidth}{!}{
\begin{tabular}{@{}l@{\hspace{3em}}l@{\hspace{3em}}
                    a@{\hspace{10pt}}c@{\hspace{2em}}
                    a@{\hspace{10pt}}c@{\hspace{2em}}
                    a@{\hspace{10pt}}c@{\hspace{2em}}
                    a@{\hspace{10pt}}c@{\hspace{2em}}
                    a@{\hspace{10pt}}c@{}}
\toprule
\multicolumn{1}{@{}c@{\hspace{3em}}}{{\textbf{Category}}} 
& \multicolumn{1}{@{}c@{\hspace{3em}}}{{\textbf{Task}}} 
& \multicolumn{6}{@{}c@{\hspace{2em}}}{\textbf{Closed Model}} 
& \multicolumn{2}{@{}c@{\hspace{2em}}}{\textbf{Open Model}} 
& \multicolumn{2}{@{}c@{\hspace{2em}}}{\textbf{Ours}}
\\
& 
                    & \multicolumn{2}{@{}c@{\hspace{2em}}}{GPT-4o}
                    & \multicolumn{2}{@{}c@{\hspace{2em}}}{GPT-4V} 
                    & \multicolumn{2}{@{}c@{\hspace{2em}}}{Gemini-1.5 Pro} 
                    & \multicolumn{2}{@{}c@{\hspace{2em}}}{LLaVA-1.6-7B} 
                    & \multicolumn{2}{@{}c@{\hspace{2em}}}{\MethodName} \\

\midrule
\multirow{1}{*}{\begin{tabular}[c]{@{}c@{}}Non-Interactive\end{tabular}}    & navigation & 0.50 & 1.00 & 1.00 & 1.00 & 1.00 & 1.00 & 0.00 & 0.00 & 1.00 & 1.00 \\
\hline
\multirow{4}{*}{\begin{tabular}[c]{@{}c@{}}Object-Interactive\end{tabular}} & shake & 0.24 & 0.27 & 0.33 & 0.34 & 0.35 & 0.24 & 0.10 & 0.18 & 0.98 & 0.91 \\
   & stir & 0.25 & 0.45 & 0.08 & 0.18 & 0.00 & 0.00 & 0.28 & 0.45 & 0.79 & 1.00 \\
   & pick and place & 0.00 & 0.00 & 0.08 & 0.29 & 0.00 & 0.00 & 0.14 & 0.43 & 1.00 & 1.00 \\
   & spread condiment & 0.20 & 1.00 & 0.15 & 0.55 & 0.00 & 0.00 & 0.00 & 0.00 & 1.00 & 0.82 \\
\hline
\multirow{3}{*}{\begin{tabular}[c]{@{}c@{}}User-Interactive\end{tabular}}    & brush hair & 0.26 & 0.48 & 0.18 & 0.58 & 0.30 & 0.23 & 0.20 & 0.45 & 0.96 & 0.74 \\
   & style hair & 0.26 & 0.85 & 0.23 & 0.46 & 0.25 & 0.42 & 0.00 & 0.00 & 1.00 & 0.77 \\
   & tidy hair & 0.17 & 0.30 & 0.38 & 0.80 & 0.40 & 0.22 & 0.18 & 0.70 & 1.00 & 0.40 \\
\hline
Average &  & 0.24 & 0.45 & 0.21 & 0.45 & 0.22 & 0.19 & 0.13 & 0.29 & \textbf{0.95} & \textbf{0.82} \\
\bottomrule
\end{tabular}}
\caption{\textbf{Comparison with SOTA off-the-shelf VLMs in \DatasetName val split.} We report precision {\small$(=TP/(TP+FP))$} and recall {\small$(=TP/(TP+FN))$} in gray and white cells, respectively. 
}\label{tab:comparison-sota-vlms-validation}
\end{table*}

\begin{table*}[t!]
\setlength{\aboverulesep}{0pt}
\setlength{\belowrulesep}{0pt}
\setlength{\extrarowheight}{.5ex}
\resizebox{\textwidth}{!}{
\begin{tabular}{@{}l@{\hspace{3em}}l@{\hspace{3em}}
                    a@{\hspace{10pt}}c@{\hspace{2em}}
                    a@{\hspace{10pt}}c@{\hspace{2em}}
                    a@{\hspace{10pt}}c@{\hspace{2em}}
                    a@{\hspace{10pt}}c@{\hspace{2em}}
                    a@{\hspace{10pt}}c@{}}
\toprule
\multicolumn{1}{@{}c@{\hspace{3em}}}{{\textbf{Category}}} 
& \multicolumn{1}{@{}c@{\hspace{3em}}}{{\textbf{Task}}} 
& \multicolumn{6}{@{}c@{\hspace{2em}}}{\textbf{Closed Model}} 
& \multicolumn{2}{@{}c@{\hspace{2em}}}{\textbf{Open Model}} 
& \multicolumn{2}{@{}c@{\hspace{2em}}}{\textbf{Ours}}
\\
& 
                    & \multicolumn{2}{@{}c@{\hspace{2em}}}{GPT-4o}
                    & \multicolumn{2}{@{}c@{\hspace{2em}}}{GPT-4V} 
                    & \multicolumn{2}{@{}c@{\hspace{2em}}}{Gemini-1.5 Pro} 
                    & \multicolumn{2}{@{}c@{\hspace{2em}}}{LLaVA-1.6-7B} 
                    & \multicolumn{2}{@{}c@{\hspace{2em}}}{\MethodName} \\

\midrule
\multirow{4}{*}{\begin{tabular}[c]{@{}c@{}}Object-Interactive\end{tabular}} & shake & 0.33 & 0.25 & 0.00 & 0.00 & 0.00 & 0.00 & 0.00 & 0.00 & 0.75 & 0.75 \\
   & stir & 0.00 & 0.00 & 0.17 & 0.50 & 0.00 & 0.00 & 0.00 & 0.00 & 0.33 & 1.00 \\
   & spread condiment & 0.20 & 0.75 & 0.17 & 0.52 & 0.15 & 0.11 & 0.11 & 0.13 & 0.98 & 0.89 \\
\hline
\multirow{3}{*}{\begin{tabular}[c]{@{}c@{}}User-Interactive\end{tabular}}    & brush hair & 0.00 & 0.00 & 0.00 & 0.00 & 0.00 & 0.00 & 0.00 & 0.00 & 1.00 & 1.00 \\
   & style hair & 0.19 & 0.44 & 0.23 & 0.32 & 0.18 & 0.36 & 0.07 & 0.20 & 1.00 & 0.96 \\
   & tidy hair & 0.23 & 0.50 & 0.42 & 0.72 & 0.43 & 0.26 & 0.28 & 0.54 & 1.00 & 0.85 \\
\hline
Average &  & 0.21 & 0.57 & 0.25 & 0.54 & 0.24 & 0.22 & 0.16 & 0.29 & \textbf{0.95} & \textbf{0.89} \\
\bottomrule
\end{tabular}}
\caption{\textbf{Comparison with SOTA off-the-shelf VLMs in \DatasetName test split.} We report precision {\small$(=TP/(TP+FP))$} and recall {\small$(=TP/(TP+FN))$} in gray and white cells, respectively. 
}\label{tab:comparison-sota-vlms-test}
\end{table*}

\appsubsection{Detailed Analysis}
\textbf{Comparison with off-the-shelf VLMs.}
\cref{tab:comparison-sota-vlms-validation} and \cref{tab:comparison-sota-vlms-test} show the performance of off-the-shelf VLMs and \MethodName for each task category. The results show that our model outperforms existing VLMs in most tasks. In the val split of \DatasetName, API-based closed VLMs show the weakest performance in `pick and place' and the strongest performance in `navigation'. In the test split of \DatasetName, closed VLMs show the weakest performance in `brush hair'. On the other hand, our model scores high precision and recall in both val and test splits, scoring $0.95$ precision and $0.89$ recall in the test split in average. Since the test split contains numerous data samples with unseen motions, tasks, and environments, the results demonstrate the generalization capability of our model. 

\begin{figure*}[ht!]{
\centering
\begin{center}
\includegraphics[width=1.0\linewidth]{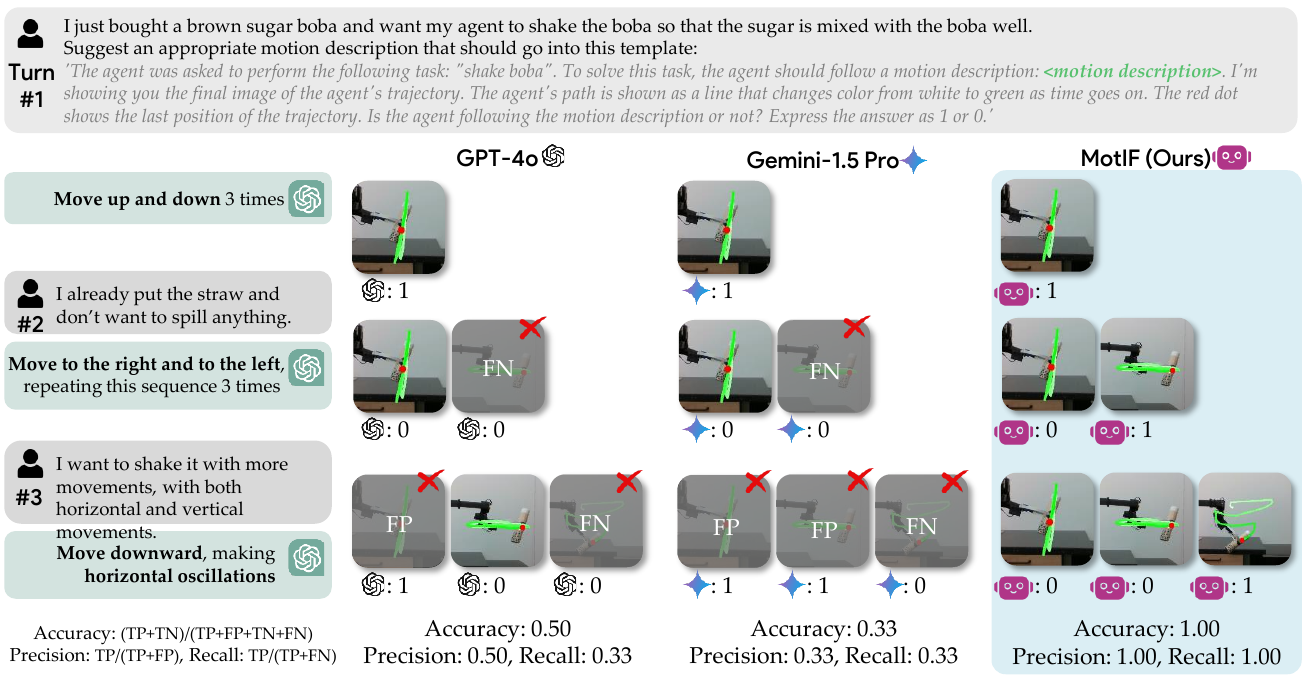}
\end{center}
\centering}
\caption{\textbf{Comparison between \MethodName and state-of-the-art closed VLMs.}
We compare three VLMs: our model, GPT-4o, and Gemini-1.5 Pro, along a conversation between a user and an LLM. The user specifies the task and the LLM generates an appropriate motion description. The performance of each VLM is measured by predicting if the robot motions align (VLM response: 1) with motion descriptions suggested from the LLM or not (VLM response: 0), where the images are not included in training our model. Comparing the accuracy, precision, and recall for each model, \MethodName shows the highest performance in all metrics.
} \label{fig:overview}
\end{figure*}

\cref{fig:overview} shows a toy scenario comparing GPT-4o, Gemini-1.5 Pro, and \MethodName, along a conversation between a user and an LLM (ChatGPT). Given a user's task specification (\textit{e.g.}, I don't want to spill anything), an LLM can generate a motion description that corresponds to the user's situation or preference. When motion description is specified from the LLM, we retrieve the best corresponding trajectory from \DatasetName, based on the judgement of a human expert. Then, we ask each VLM to determine whether the robot's motion corresponds to the motion description or not. At each turn, we ask the VLMs to match the motion description proposed from the LLM with the current and all previously retrieved motions. After three turns of conversation, we calculate the performance of motion discrimination using the VLMs. Results in \cref{fig:overview} show that our model achieves the highest accuracy, precision, and recall, successfully understanding robotic motions in all cases. 

Another interesting aspect of off-the-shelf VLMs is that they can refuse to answer or output a neutral response. For instance, Gemini-1.5 Pro often outputs ``The answer is: **Cannot be determined**.'', ``We cannot tell, so we cannot provide a 1 or 0 answer.'', or ``it's impossible to tell if the agent followed the instruction''. GPT-4V frequently outputs ``I'm sorry, I can't provide assistance with that request.'' or ``I'm sorry, but I cannot provide assistance with tasks or interpret content that includes tracking or analyzing the movements''.

\begin{table*}[ht!]
\setlength{\aboverulesep}{0pt}
\setlength{\belowrulesep}{0pt}
\setlength{\extrarowheight}{.5ex}
\renewcommand{\arraystretch}{1.1}
\resizebox{\textwidth}{!}{
\begin{tabular}{@{}c@{\hspace{3em}}c@{\hspace{3em}}c@{\hspace{3em}}
                    a@{\hspace{10pt}}c@{\hspace{2em}}
                    a@{\hspace{10pt}}c@{\hspace{2em}}
                    }
\toprule
\multicolumn{1}{@{}c@{\hspace{3em}}}{{\textbf{Model}}} 
& 
\multicolumn{2}{@{}c@{\hspace{2em}}}{\textbf{Training Data}} 
& \multicolumn{2}{@{}c@{\hspace{2em}}}{\textbf{Val Performance}} 
& \multicolumn{2}{@{}c@{\hspace{2em}}}{\textbf{Test Performance}} 
\\
                    & \multicolumn{1}{@{}c@{\hspace{2em}}}{\# human demos}
                    & \multicolumn{1}{@{}c@{\hspace{2em}}}{\# robot demos} 
                    & \multicolumn{1}{@{}c@{\hspace{2em}}}{Precision} 
                    & \multicolumn{1}{@{}c@{\hspace{2em}}}{Recall} 
                    & \multicolumn{1}{@{}c@{\hspace{2em}}}{Precision} 
                    & \multicolumn{1}{@{}c@{\hspace{2em}}}{Recall}
                    
\\
\midrule
Human Only & 653 & 0 & 0.63 & 0.50 & 0.92 & 0.33 \\
\midrule
\multirow{3}{*}{\begin{tabular}[c]{@{}c@{}}Robot Only\end{tabular}} & 653 & 20 & 0.55 & 0.26 & 0.58 & 0.38 \\
 & 653 & 50 & 0.68 & 0.87 & 0.53 & 0.85 \\
 & 653 & 100 & 0.83 & \textbf{0.89} & 0.89 & 0.83 \\
\midrule
\multirow{3}{*}{\begin{tabular}[c]{@{}c@{}}Human+Robot\\ Co-trained\end{tabular}} & 653 & 20 & 0.77 & 0.68 & 0.95 & 0.83 \\
 & 653 & 50 & 0.92 & \textbf{0.89} & \textbf{0.96} & 0.84 \\
 & 653 & 100 & \textbf{0.95} & 0.82 & 0.95 & \textbf{0.89} \\
\bottomrule
\end{tabular}}
\caption{\textbf{Co-training Performance in \DatasetName Val and Test splits.} We report precision {\small$(=TP/(TP+FP))$} and recall {\small$(=TP/(TP+FN))$} in gray and white cells, respectively. 
}\label{tab:co-training-performance}
\end{table*}

\textbf{Co-training on Human and Robot Datasets.}
\cref{tab:co-training-performance} and \cref{fig:co-training-graph} show detailed results of how co-training on human and robot data improves performance and boosts positive transfer. To focus on the effect of co-training, we fix the visual motion representation as single point tracking. Adding just 20 robot demonstrations to all human data, the recall score significantly improves by $169.7\%$ in the test split.

\begin{table*}[t!]
\setlength{\aboverulesep}{0pt}
\setlength{\belowrulesep}{0pt}
\setlength{\extrarowheight}{.5ex}
\renewcommand{\arraystretch}{1.1}
\resizebox{\textwidth}{!}{
\begin{tabular}{@{}c@{\hspace{3em}}c@{\hspace{3em}}c@{\hspace{3em}}
                    a@{\hspace{10pt}}c@{\hspace{2em}}
                    a@{\hspace{10pt}}c@{\hspace{2em}}
                    }
\toprule
\multicolumn{3}{@{}c@{\hspace{2em}}}{\textbf{Visual Motion Representation}} 
& \multicolumn{2}{@{}c@{\hspace{2em}}}{\textbf{Val Performance}} 
& \multicolumn{2}{@{}c@{\hspace{2em}}}{\textbf{Test Performance}} 
\\
Type
                    & \multicolumn{1}{@{}c@{\hspace{2em}}}{Trajectory Drawing}
                    & \multicolumn{1}{@{}c@{\hspace{2em}}}{\# frames} 
                    & \multicolumn{1}{@{}c@{\hspace{2em}}}{Precision} 
                    & \multicolumn{1}{@{}c@{\hspace{2em}}}{Recall} 
                    & \multicolumn{1}{@{}c@{\hspace{2em}}}{Precision} 
                    & \multicolumn{1}{@{}c@{\hspace{2em}}}{Recall}
                    
\\
\midrule
Raw Image & \xmark & 1 & 0.92 & 0.76 & \textbf{0.97} & 0.79 \\
\midrule
\multirow{3}{*}{\begin{tabular}[c]{@{}c@{}}Multi-Frame\\ Storyboard\end{tabular}} & \xmark & 2 & 0.89 & 0.81 & 0.92 & 0.86 \\
 & \xmark & 4 & 0.87 & 0.87 & 0.96 & 0.78 \\
 & \xmark & 9 & 0.87 & 0.87 & 0.95 & 0.83 \\
\midrule
Optical Flow & \checkmark & 1 & \textbf{0.97} & \textbf{0.88} & 0.95 & 0.74 \\
\midrule
\MethodName (Ours) & \checkmark & 1 & 0.95 & 0.82 & 0.95 & \textbf{0.89} \\
\bottomrule
\end{tabular}}
\caption{\textbf{Comparison of Visual Motion Representations in \DatasetName Val and Test splits.} We report precision {\small$(=TP/(TP+FP))$} and recall {\small$(=TP/(TP+FN))$} in gray and white cells, respectively. 
}\label{tab:comparison-representations}
\end{table*}

\textbf{Visual Motion Representations.}
\cref{tab:comparison-representations} shows detailed performance values of models trained with different visual motion representations. Using the raw last frame shows the highest precision in the test split, while the recall score is $11.2\%$ lower than using the proposed motion representation, drawing the trajectory of a single keypoint. This implies that without the information of the trajectory, the VLM can easily output false negatives. Bringing insights from GPT-4o generating storyboard when analyzing videos, we also demonstrate the performance using multi-frame storyboards with 2, 4, and 9 keyframes. Interestingly, higher number of frames does not consistently improve performance, which might be due to the lower resolution of images when multiple images are concatenated. 

\clearpage

\noindent
\begin{minipage}{\linewidth}
\appsubsection{Visualization of \MethodName Outputs}
In this section, we visualize four trajectories in the val and test splits of \DatasetName. For each trajectory, we analyze the output of \MethodName given four different motion descriptions.
\vspace{1em}

\cref{tab:trajviz1} visualizes the $106^{th}$ trajectory in the val split of \MethodName. While motion description 1 is included in the training data, our VLM generalizes to understanding unseen motion descriptions (2, 3, and 4). This may be due to the effect of having diverse human demonstrations on grounded robot motions, since there are similar motion descriptions in human data such as ``move downward, getting closer to the laptop, and then move to the left''.

\cref{tab:trajviz2} shows an example of our model understanding a robotic motion from an unseen camera viewpoint. Despite the change of the camera height and angle, our model effectively understands a paraphrased instruction (motion description 2) and an instruction with a small tweak on the core directional component (motion description 4).

\cref{tab:trajviz3} demonstrates our model understands robotic motions even when the interacting object is changed. While parmesan cheese is the only object for this task in the training data, \MethodName does accurately determine ground truth and paraphrased motion descriptions align with the robot's motion. 

\cref{tab:trajviz4} shows a trajectory of a robot delivering lemonade to the door, while avoiding moving over the manhole. This example demonstrates the grounded motion understanding capability of our model such as moving over or making a detour to avoid a specific instance (e.g., manhole) in the environment.
\vspace{-10cm}
\end{minipage}

\newpage
\begin{table}[t!]
\setlength{\aboverulesep}{0pt}
\setlength{\belowrulesep}{0pt}
\setlength{\extrarowheight}{.1ex}
\resizebox{\textwidth}{!}{
\begin{tabular}{lcc}
\multicolumn{3}{c}{\parbox[c]{\linewidth}{\includegraphics[width=1\linewidth]{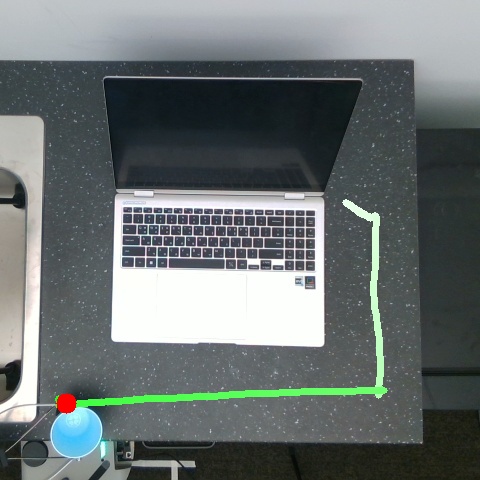}}} \vspace{10pt}\\
\toprule
\textbf{Task Instruction: pick up the cup and} & \textbf{\MethodName} & \textbf{Prediction} \\
\textbf{place it to the lower left of the laptop} & \textbf{Output} & \textbf{Correctness} \\
\midrule
Motion Description 1: move downward, & 1 & \greencheck \\
then move to the left &  &  \\\hline
Motion Description 2: move farther from the  & 1 & \greencheck \\
laptop, move downward, then move to the left &  &  \\\hline
Motion Description 3: move downward and  & 0 & \greencheck \\
to the left, passing over the laptop &  &  \\\hline
Motion Description 4: move over the laptop & 0 & \greencheck \\
\bottomrule
\end{tabular}}
\caption{\textbf{Trajectory Visualization of a Grounded Motion.} For the $106^{th}$ robot motion in the validation split of \DatasetName, we get the output of \MethodName on 4 different motion descriptions. The results show that our model effectively understands how the robot's motion is grounded in the environment, especially considering the spatial relationship with the laptop.}\label{tab:trajviz1}
\end{table}

\begin{table}[ht!]
\setlength{\aboverulesep}{0pt}
\setlength{\belowrulesep}{0pt}
\setlength{\extrarowheight}{.1ex}
\resizebox{\textwidth}{!}{
\begin{tabular}{lcc}
\multicolumn{3}{c}{\parbox[c]{1\linewidth}{\includegraphics[width=1\linewidth]{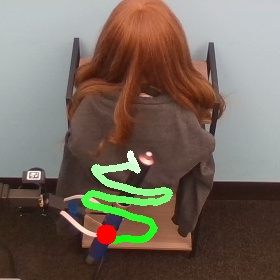}}} \vspace{10pt}\\
\toprule
\textbf{Task Instruction: curl hair} & \textbf{\MethodName} & \textbf{Prediction} \\
 & \textbf{Output} & \textbf{Correctness} \\\midrule
Motion Description 1: move downward,  & 1 & \greencheck \\
while making horizontal oscillations &  &  \\\hline
Motion Description 2: move downward,  & 1 & \greencheck \\
while making side-to-side movements &  &  \\\hline
Motion Description 3: move downward & 0 & \greencheck \\\hline
Motion Description 4: move downward,  & 0 & \greencheck \\
while making vertical oscillations &  &  \\\hline
\bottomrule
\end{tabular}}
\caption{\textbf{Trajectory Visualization of a Motion in an Unseen Camera Viewpoint.} For the $14^{th}$ robot motion in the test split of \DatasetName, we get the output of \MethodName on 4 different motion descriptions. Our model effectively understands the robot's motion in an unseen camera viewpoint.}\label{tab:trajviz2}
\end{table}

\begin{table}[ht!]
\setlength{\aboverulesep}{0pt}
\setlength{\belowrulesep}{0pt}
\setlength{\extrarowheight}{.1ex}
\resizebox{\textwidth}{!}{
\begin{tabular}{lcc}
\multicolumn{3}{c}{\parbox[c]{1\linewidth}{\includegraphics[width=1\linewidth]{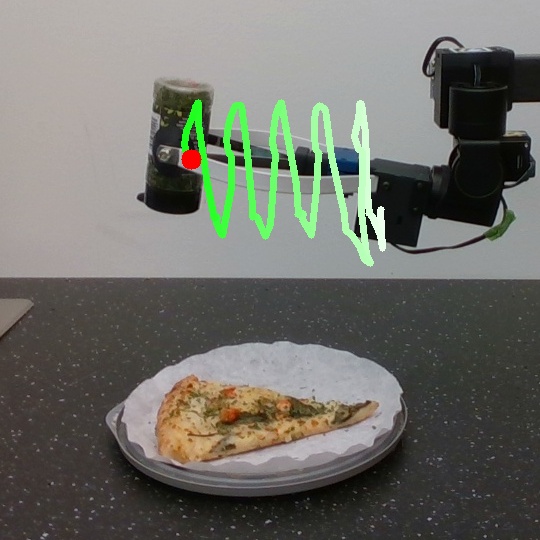}}} \vspace{10pt}\\
\toprule
\textbf{Task Instruction: sprinkle parsley on pizza} & \textbf{\MethodName } & \textbf{Prediction } \\ 
 & Output & Correctness \\ \midrule
Motion Description 1: move to the left, while  & 1 & \greencheck \\
making vertical oscillations and alternating rotations &  &  \\\hline
Motion Description 2: move to the left, while  & 1 & \greencheck \\
making vertical oscillations &  &  \\\hline
Motion Description 3: move to the left, while  & 1 & \greencheck \\
making vertical shaking movements &  &  \\\hline
Motion Description 4: move to the left in a  & 0 & \greencheck \\
straight line &  &  \\
\bottomrule
\end{tabular}}
\caption{\textbf{Trajectory Visualization of a Motion with an Unseen Object.} For the $166^{th}$ robot motion in the test split of \DatasetName, we get the output of \MethodName on 4 different motion descriptions. While the training data only contains motions of spreading condiments on pizza with parmesan cheese, our model effectively understands the robot's motion with an unseen object, parsley.}\label{tab:trajviz3}
\end{table}

\begin{table}[ht!]
\setlength{\aboverulesep}{0pt}
\setlength{\belowrulesep}{0pt}
\setlength{\extrarowheight}{.1ex}
\resizebox{\textwidth}{!}{
\begin{tabular}{lcc}
\multicolumn{3}{c}{\parbox[c]{1\linewidth}{\includegraphics[width=1\linewidth]{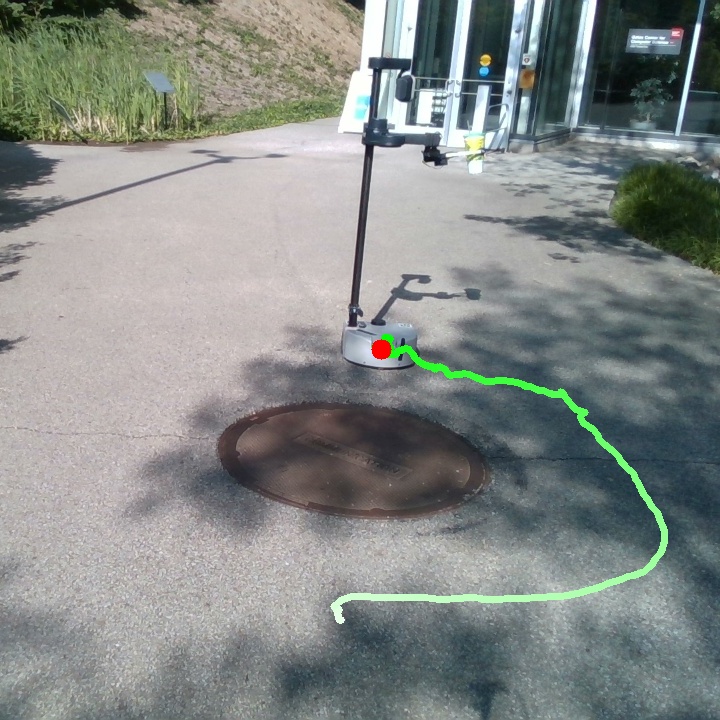}}} \vspace{10pt}\\
\toprule
\textbf{Task Instruction: deliver lemonade} & \textbf{\MethodName } & \textbf{Prediction } \\ 
 & Output & Correctness \\\midrule
Motion Description 1: make a detour  & 1 & \greencheck \\
to the right of the manhole &  &  \\\hline
Motion Description 2: move forward,  & 1 & \greencheck \\
making a detour to the right of the manhole &  &  \\\hline
Motion Description 3: move forward  & 0 & \greencheck \\
in the shortest path &  &  \\\hline
Motion Description 4: move forward  & 0 & \greencheck \\
in a straight line, moving over the manhole &  &  \\
\bottomrule
\end{tabular}}
\caption{\textbf{Trajectory Visualization of an Unseen Grounded Motion.} For the $181^{th}$ robot motion in the val split of \DatasetName, we get the output of \MethodName on 4 different motion descriptions. The results show that our model effectively understands the semantic grounding of the robot's motion in a navigation task.}\label{tab:trajviz4}
\end{table}

\appsubsection{Qualitative Analysis on Off-the-shelf VLMs}

\begin{figure}[ht!]{
\centering
  \begin{center}
\includegraphics[width=1.0\textwidth]{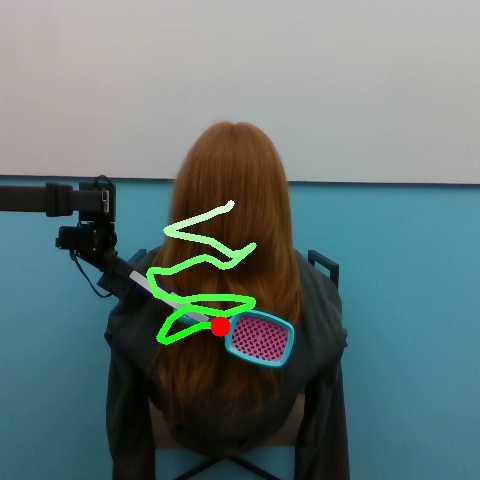}
  \end{center}
  \caption{\small{\textbf{Example Trajectory of a robot brushing hair.} 
  The robot is brushing the hair by moving downward, while making horizontal oscillations. }\vspace{-0pt}}
  \label{fig:example-traj-brush-hair}
  }
\end{figure}

To check the motion understanding capabilities of off-the-shelf VLMs, we perform a toy experiment on asking several questions to the VLMs, given a video or an image of the robot's motion brushing hair (see \cref{fig:example-traj-brush-hair}). The results in this section show that off-the-shelf VLMs such as GPT-4V and GPT-4o fail to describe the robot's motion and its semantic grounding in the environment.

\cref{prompt:raw-video-output-1} shows the response of GPT-4V and GPT-4o, given the robot's motion video, without any trajectory overlaid, and a simple text prompt `Describe the robot's motion'. While GPT-4V fails to analyze the video, GPT-4o generates a long detailed response explaining the robot's motion. Specifically, GPT-4o first extracts keyframes with its internal algorithm and creates a storyboard of the keyframes. Then, it detects the trajectory of the robot's end effector. Although the model's approach is logically valid, the model eventually describes the robot's motion as "The robot's end effector moves in a non-linear path with variations in both the x and y directions, suggesting a complex motion.", which implies neither a specific shape of the motion or how it is grounded in the scene.

\begin{figure*}[ht!]
\centering
\begin{minted}[mathescape, breaklines, fontsize=\footnotesize, fontsize=\small, numbersep=8pt, escapeinside=||, breaksymbolleft={}, breaksymbolright={}, bgcolor=LightGray]{text}
# Question
Describe the robot's motion.
\end{minted}

\begin{minted}[mathescape, breaklines, fontsize=\footnotesize, fontsize=\small, numbersep=8pt, escapeinside=||, breaksymbolleft={}, breaksymbolright={}, bgcolor=LightGray]{text}
# GPT-4V Answer
It seems you've uploaded a video file related to the robot's motion. Let's analyze the content of this video to describe the motion in detail. I'll review the video and provide a description of the robot's movements. Please hold on for a moment.

The video is ready for viewing. Please watch the robot's motion in the video above. After viewing, I can help with any specific analysis or description you need about the robot's movements. If you have specific aspects of the motion you're interested in, such as speed, trajectory, or interaction with objects, let me know!
\end{minted}

\begin{minted}[mathescape, breaklines, fontsize=\footnotesize, fontsize=\small, numbersep=8pt, escapeinside=||, breaksymbolleft={}, breaksymbolright={}, bgcolor=LightGray]{text}

# GPT-4o Answer
The first frame of the video shows the side view of a robot performing a task. To describe the robot's motion, I'll need to analyze the frames in the video to identify the movement pattern. I'll extract and analyze key frames to provide a detailed description of the motion.

Let's proceed by extracting key frames and tracking the movement of the robot's end effector.

The extracted key frames from the video show the robot's motion over time. To provide a detailed description of the robot's motion, let's analyze these frames:

The robot's end effector appears to be moving in a specific path throughout the video.
The trajectory includes several distinct positions and orientations at different times.
To describe the motion more precisely, I will track the movement of the robot's end effector across these frames. This will help in visualizing and understanding the pattern of movement.

The tracked positions of the robot's end effector across the frames indicate its motion pattern. Here is the detailed description of the robot's motion:

Initial Position: The robot starts with the end effector near the top left quadrant of the frame.

Motion Path: The end effector follows a somewhat curved path, moving diagonally towards the bottom right quadrant.

Key Positions:
Frame 1: (146, 166)\n Frame 2: (174, 173)\n Frame 3: (303, 221)\n Frame 4: (260, 236)
Frame 5: (296, 245)\n Frame 6: (265, 225)\n Frame 7: (300, 266)\n Frame 8: (245, 282)
Frame 9: (159, 258)\n Frame 10: (289, 319)\n Frame 11: (262, 353)\n Frame 12: (318, 336)

Overall Trajectory: The robot's end effector moves in a non-linear path with variations in both the x and y directions, suggesting a complex motion.
The movement is likely part of a task requiring precise positioning and orientation changes. If you have any specific aspects of the motion or additional details you'd like to focus on, please let me know!

\end{minted}
\caption{Example prompt and answer where GPT-4o is asked to describe the robot's motion. The model is given a video of the robot's motion with the question prompt. The video does not include any trajectory visualization. Given the video, the model fails to describe the semantic meaning of the robot's motion.}
\label{prompt:raw-video-output-1}
\end{figure*}

We also check if off-the-shelf VLMs can better understand robotic motions given the proposed visual motion representation, by overlaying the trajectory of a single keypoint in the image. \cref{prompt:traj-video-output} shows an example of passing a video with the robot's trajectory visualized. While GPT-4o outputs a response with detailed explanation, the described motion with multiple downward strokes does not match with the actual motion as shown in \cref{fig:example-traj-brush-hair}, which is moving downward while making horizontal oscillations. This implies that existing VLMs might rely on the language prior, because making multiple downward strokes is a common brushing technique in general.

\begin{figure*}[t!]
\centering
\begin{minted}[mathescape, breaklines, fontsize=\footnotesize, fontsize=\small, numbersep=8pt, escapeinside=||, breaksymbolleft={}, breaksymbolright={}, bgcolor=LightGray]{text}
# Question
Describe the robots motion. The task is 'brushing hair'. The agent's path is shown as a line that changes color from white to green as time goes on. THe red dot shows the last position of the trajectory.
\end{minted}

\begin{minted}[mathescape, breaklines, fontsize=\footnotesize, fontsize=\small, numbersep=8pt, escapeinside=||, breaksymbolleft={}, breaksymbolright={}, bgcolor=LightGray]{text}
# GPT-4o Answer
I've loaded the video of the robot's motion for the task of brushing hair. To provide a detailed description of the motion, I will analyze the video and describe the path and color change over time, along with the final position indicated by the red dot.

Let's start by extracting key frames and analyzing the motion trajectory.
<12x1 storyboard displayed>

The key frames extracted from the video provide a visual representation of the robot's motion for the task of brushing hair. Here's a detailed description of the motion:

Initial Position: The robot starts with the brush positioned near the top of the hair, as indicated by the white color at the beginning of the trajectory.

Motion Path: The robot moves the brush in a downward stroke, following a smooth and continuous path. The color of the trajectory gradually changes from white to green, indicating the progression of time.

Path Continuation: The robot completes the first downward stroke and moves the brush back up to the starting position, preparing for another stroke. The color of the line continues to change from white to green, showing the ongoing motion.

Red Dot: The red dot at the end of the trajectory indicates the last position of the brush at the end of the video. This dot signifies the completion of the brushing motion.

Overall, the robot's motion is characterized by a repetitive up-and-down brushing movement, with the color-changing line visualizing the time progression of the motion. The red dot marks the final position of the brush.

\end{minted}
\caption{Example prompt and answer where GPT-4o is asked to describe the robot's motion. The model is given a video of the robot's motion with the robot's trajectory overlaid, and the question prompt.}
\label{prompt:traj-video-output}
\end{figure*}

\clearpage
\section{Examples of Using \MethodName}
\label{app-sec:implications-motif}

\begin{figure*}[ht!]{
\centering
\begin{center}
\includegraphics[width=0.7\linewidth]{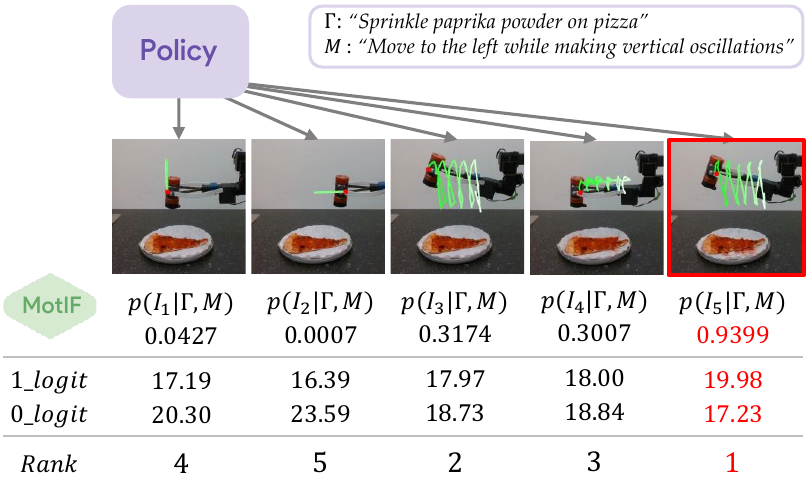}
\end{center}
\centering}
\caption{\textbf{Ranking Trajectories - Case 1. Unseen Object.} We can use the probability of the output token being "1" or the logits for tokens "1" and "0" as a criterion to rank the trajectories. The bottom row shows the rank determined by the probability in the first row.
} \label{fig:ranking-traj-1}
\end{figure*}

\begin{figure*}[ht!]{
\centering
\begin{center}
\includegraphics[width=0.85\linewidth]{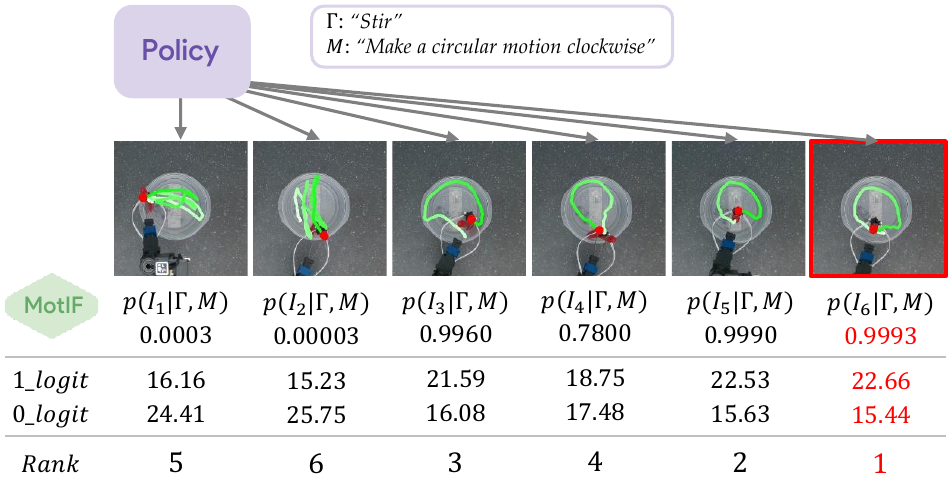}
\end{center}
\centering}
\caption{\textbf{Ranking Trajectories - Case 2. Unseen Motions.} We can use the probability of the output token being "1" or the logits for tokens "1" and "0" as a criterion to rank the trajectories. The bottom row shows the rank determined by the probability in the first row.
} \label{fig:ranking-traj-2}
\end{figure*}

In this section, we provide a detailed examination of the qualitative results demonstrating the utility of \MethodName in evaluating policy-generated trajectories. Specifically, we focus on how \MethodName acts as a critical component in assessing whether the motions of a robot are in alignment with the given task instructions and motion descriptions. This capability is essential for determining appropriate moments to terminate episodes or to offer corrective feedback to the policy module responsible for trajectory generation. We illustrate various applications of \MethodName, including its role as a success detector, providing correction feedback, terminating or adapting robot policies. Additionally, we discuss how \MethodName can be used to rank trajectories, as depicted in \cref{fig:ranking-traj-1} and \cref{fig:ranking-traj-2}. These examples underscore the method's potential in various robotic applications.

\appsubsection{Refining and Terminating Robot Planning}
\textbf{Policy.}
Bringing an insight from code-as-policies~\cite{liang2023codeaspolicies}, we use an LLM (ChatGPT) to generate the agent policy as a code. We use the prompts in \cref{prompt:code-as-policies} to generate a function that outputs the sequence of end effector states. We make the functions to be parameterized, so that we can utilize the output of \MethodName as a correction feedback to adjust the parameters. In \cref{tab:refining=terminating-robot-planning}, we show an example of refining a policy to generate a robot's motion corresponding to "move to the left while making vertical oscillations".

\begin{table*}[ht!]\vspace{30pt}
\setlength{\aboverulesep}{0pt}
\setlength{\belowrulesep}{0pt}
\setlength{\extrarowheight}{.1ex}
\resizebox{\textwidth}{!}{
\begin{tabular}{lclcc}
\toprule
\textbf{Turn} & \textbf{Trajectory} & \textbf{Policy} & \textbf{\MethodName output} & \textbf{Prob. of "1"} \\
\midrule
1 & \multicolumn{1}{c}{\parbox[c]{0.4\linewidth}{\includegraphics[width=0.8\linewidth]{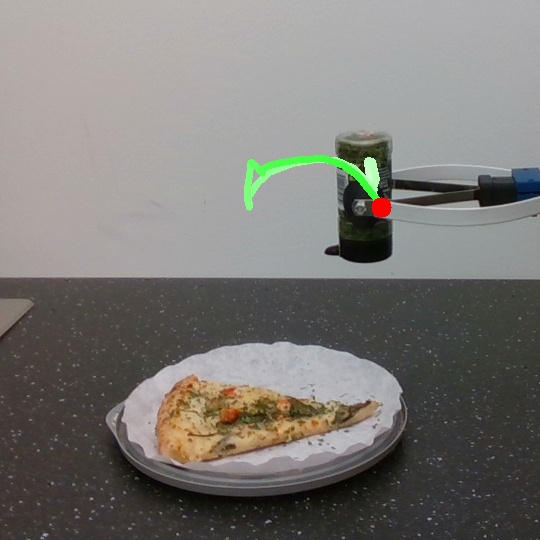}}} & horizontal$\_$shaking$(2, 0.1, (0.1, 1.0), 0.15, 2, roll_angle=3.14)$ & 0 & 0.392\\ 
\midrule
2 & \multicolumn{1}{c}{\parbox[c]{0.4\linewidth}{\includegraphics[width=0.8\linewidth]{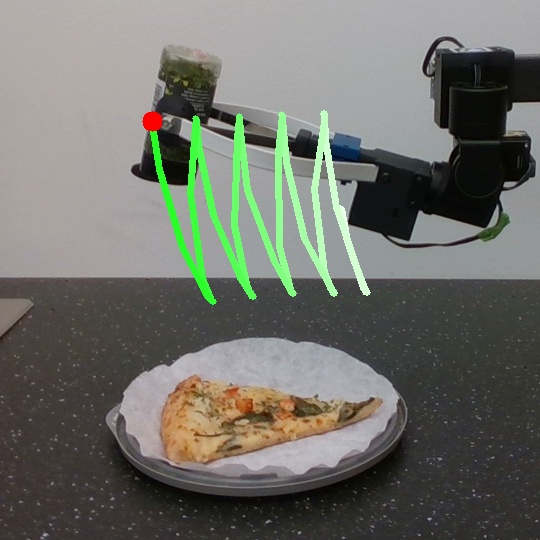}}} & vertical$\_$shaking$(2, 0.1, (0.2, 1.0), 0.1, 4)$ & 1 & 0.999 \\
\bottomrule
\end{tabular}}
\caption{\textbf{Refining and Terminating Policy with \MethodName Output as Feedback.} 
For task instruction "sprinkle parsley on pizza" and motion description "move to the left while making vertical oscillations", an LLM suggests to use horizontal$\_$shaking function with wrist roll rotations in the first turn. Using the output of \MethodName or the probability of output "1" as a signal, we can terminate when the output is "1" or the probability is greater than a threshold, and refine the policy otherwise.
}\label{tab:refining=terminating-robot-planning}
\end{table*}

\clearpage
\appsubsection{Ranking Trajectories with \MethodName}

Another usage of \MethodName can be ranking trajectories, as shown in \cref{fig:ranking-traj-1} and \cref{fig:ranking-traj-2}. To rank trajectories, we can use the probability of the output token being "1" or the logits for tokens "1" and "0" as criteria. While these different criteria make a consensus on choosing the best trajectory in both examples (the rightmost trajectories in \cref{fig:ranking-traj-1} and \cref{fig:ranking-traj-2}), the ranking of the other trajectories change depending on the criteria. For instance, in \cref{fig:ranking-traj-1}, while the rank from the first to the last trajectories is $(4, 5, 2, 3, 1)$ using the probability, it becomes $(4, 5, 3, 2, 1)$ using the logit for token "1". Despite a small difference on the rankings, the two leftmost trajectories always rank the lowest. This is a desirable result since those trajectories only show a simple translation motion, without any oscillations. 

\cref{fig:ranking-traj-2} shows how we can reduce human effort in choosing high-quality demonstrations. In this example, robot motions get higher scores as the shape becomes closer to a perfect circle. Also, the two leftmost trajectories with repeated vertical or horizontal movements are accurately determined as failed trajectories.

\begin{figure*}[t]
\centering
\begin{minted}[mathescape, breaklines, fontsize=\footnotesize, fontsize=\small, numbersep=8pt, escapeinside=||, breaksymbolleft={}, breaksymbolright={}, bgcolor=LightGray]{text}
# Generate motion primitive functions
I want to generate videos of a 2d agent moving by following a given path with specific shapes. for instance, an agent can follow a linear line from point (0,0) to (1,1), while another agent can follow a sine wave from (0,0.5) to (1,0.5). Set the range of x,y coordinates from 0 to 1. Write down a python code to generate N different trajectories and save those as .mp4 videos and also .jpg images. Also, save the information of trajectories in a .json file, including the shape type of the path, starting point, and ending point.

\end{minted}

\begin{minted}[mathescape, breaklines, fontsize=\footnotesize, fontsize=\small, numbersep=8pt, escapeinside=||, breaksymbolleft={}, breaksymbolright={}, bgcolor=LightGray]{text}

# Generate a robot's motion moving from point A to point B in a straight line
def draw_line_position_mode(n, time_dt=1.5, start=(0,0.5), end=(0.5,1.0)):
    x = np.linspace(start[0], end[0], n, endpoint=True)
    y = np.linspace(start[1], end[1], n, endpoint=True)
    mat = np.c_[x, y]
    episode_info = {"type": "straight_line", "n": n, "time_dt": time_dt, "start": start, "end": end}
    return mat, episode_info

# Generate a robot's motion moving from point A to point B, while making horizontal oscillations
def horizontal_shaking(n, time_dt=1.5, start=(0,0.5), amplitude=0.1, frequency=2, roll_angle=0):
    start_point = start
    end_point = (start[0] + amplitude, start[1])
    x = np.linspace(start_point[0], end_point[0], n, endpoint=True)
    y = np.linspace(start_point[1], end_point[1], n, endpoint=True)
    mat = np.c_[x, y]
    horizontal_shaking_mat = []
    for i in range(frequency):
        if i % 2 == 0:
            horizontal_shaking_mat.append(mat)
        else:
            horizontal_shaking_mat.append(mat[::-1])
    # add wrist_roll for vibration
    wrist_roll = np.array([roll_angle, 0.0] * (n*frequency))
    episode_info = {"type": "horizontal_shaking", "n": n, "time_dt": time_dt, "start": start, "amplitude": amplitude, "frequency": frequency, "wrist_roll": wrist_roll}
    return np.concatenate(horizontal_shaking_mat), episode_info

# Generate a robot's motion moving from point A to point B, while making vertical oscillations
def vertical_shaking(n, time_dt=1.5, start=(0,0.5), amplitude=0.1, frequency=2, roll_angle=0):
    start_point = start
    end_point = (start[0], start[1] + amplitude)
    x = np.linspace(start_point[0], end_point[0], n, endpoint=True)
    y = np.linspace(start_point[1], end_point[1], n, endpoint=True)
    mat = np.c_[x, y]
    vertical_shaking_mat = []
    for i in range(frequency):
        if i % 2 == 0:
            vertical_shaking_mat.append(mat)
        else:
            vertical_shaking_mat.append(mat[::-1])
    # add wrist_roll for vibration
    wrist_roll = np.array([roll_angle, 0.0] * (n*frequency))
    episode_info = {"type": "vertical_shaking", "n": n, "time_dt": time_dt, "start": start, "amplitude": amplitude, "frequency": frequency, "wrist_roll": wrist_roll}
    return np.concatenate(vertical_shaking_mat), episode_info

\end{minted}
\caption{Example prompt and generated policies when ChatGPT is asked to generate policies as codes.}
\label{prompt:code-as-policies}
\end{figure*}

\end{document}